\documentclass{article}

\usepackage[final]{neurips_2023}
\usepackage{felix}

\begin{document}
\title{Replicability in Reinforcement Learning\thanks{Authors are listed alphabetically.}}

% The \author macro works with any number of authors. There are two commands
% used to separate the names and addresses of multiple authors: \And and \AND.
%
% Using \And between authors leaves it to LaTeX to determine where to break the
% lines. Using \AND forces a line break at that point. So, if LaTeX puts 3 of 4
% authors names on the first line, and the last on the second line, try using
% \AND instead of \And before the third author name.

\author{%
  Amin Karbasi \\
  Yale University, Google Research \\
  \texttt{amin.karbasi@yale.edu} \\
  \And
  Grigoris Velegkas \\
  Yale University \\
  \texttt{grigoris.velegkas@yale.edu} \\
  \And
  Lin F. Yang \\
  UCLA \\
  \texttt{linyang@ee.ucla.edu}
  \And
  Felix Zhou \\
  Yale University \\
  \texttt{felix.zhou@yale.edu}
}

\maketitle

\begin{abstract}
  We initiate the mathematical study of replicability as an 
  algorithmic property in the context of reinforcement learning (RL).
  We focus on the fundamental setting of discounted tabular MDPs with access to a \emph{generative model}.
  Inspired by \citet{impagliazzo2022reproducibility}, we say that an RL algorithm is replicable if,
  with high probability,
  it outputs the \emph{exact} same policy
  after two executions on i.i.d. samples drawn from the generator
  when its \emph{internal} randomness
  is the same.
  We first provide 
  an efficient $\rho$-replicable algorithm for $(\varepsilon, \delta)$-optimal policy estimation
  with sample and time complexity $\widetilde O\left(\frac{N^3\cdot\log(1/\delta)}{(1-\gamma)^5\cdot\varepsilon^2\cdot\rho^2}\right)$,
  where $N$ is the number of state-action pairs.
  Next,
  for the subclass of deterministic algorithms,
  we provide a lower bound of order $\Omega\left(\frac{N^3}{(1-\gamma)^3\cdot\varepsilon^2\cdot\rho^2}\right)$.
  Then, we study a relaxed version of replicability proposed
  by \citet{kalavasis2023statistical} called \emph{$\TV$ indistinguishability}.
  \iffalse
  We design a $\rho$-$\TV$ indistinguishable variant 
  of the Gaussian mechanism for multiple
  statistical query estimation that returns $(\varepsilon,
  \delta)$-accurate estimates using $\widetilde O\left( \frac{d^2 \log(1/\delta)}{\varepsilon^2\rho^2 } \right)$ samples in total, where $d$ is the number of queries. Using that oracle, w
  \fi
  We design a computationally efficient TV indistinguishable algorithm for policy estimation
  whose sample complexity is $\widetilde O\left(\frac{N^2\cdot\log(1/\delta)}{(1-\gamma)^5\cdot\varepsilon^2\cdot\rho^2}\right)$.
  At the cost of $\exp(N)$ running time,
  we transform these TV indistinguishable algorithms to $\rho$-replicable ones without increasing their sample complexity.
  Finally,
  we introduce the notion of \emph{approximate}-replicability
  where we only require that two outputted policies are close
  under an appropriate statistical divergence (e.g., Renyi)
  and show an improved sample complexity of $\widetilde O\left(\frac{N\cdot\log(1/\delta)}{(1-\gamma)^5\cdot\varepsilon^2\cdot\rho^2}\right)$.
\end{abstract}

\section{Introduction}
When designing a reinforcement learning (RL) algorithm, how can one ensure that when it is executed
twice in the same environment its outcome will be the same? In this work, our goal
is to design RL algorithms with \emph{provable} replicability guarantees.
The lack of replicability in 
scientific research, which the community also refers to as the \emph{reproducibility crisis},
has been a major recent concern. This can be witnessed by an article that appeared in Nature
\citep{baker20161}: Among the 1,500 scientists who participated in a survey, 70\% of them
could not replicate other researchers' findings and more shockingly,
50\% of them could
not even reproduce their own results. Unfortunately, due to the exponential increase
in the volume of Machine Learning (ML) papers that are being published each year, the ML
community has also observed an alarming increase in the lack of reproducibility. As a result, major ML 
conferences such as NeurIPS and ICLR have established ``reproducibility challenges'' in which
researchers are encouraged to replicate the findings of their colleagues \citep{pineau2019iclr,pineau2021improving}.

Recently, RL algorithms have been a crucial component
of many ML systems that are being deployed in various application
domains. These include but are not limited to, competing with humans in games \citep{mnih2013playing, silver2017mastering, vinyals2019grandmaster, meta2022human},
creating self-driving cars \citep{kiran2021deep},
designing recommendation systems \citep{afsar2022reinforcement},
providing e-healthcare services \citep{yu2021reinforcement},
and training Large Language Models (LLMs) \citep{ouyang2022training}.
In order to ensure replicability across these systems,
an important first step is to develop replicable RL algorithms. 
To the best of our knowledge, 
replicability in the context of RL has not received a formal mathematical treatment.
We initiate this effort by focusing on 
\emph{infinite horizon, tabular} RL with a \emph{generative model}.
The generative model was first studied by \citet{kearns1998finite} in order to understand the statistical complexity of long-term planning without the complication of exploration.
The crucial difference between this setting and Dynamic Programming (DP) \citep{bertsekas1976dynamic} is that
the agent needs to first obtain information about the world
before computing a \emph{policy} through some optimization process.
Thus, the main question is to understand
the number of samples required  to estimate a near-optimal policy.
\iffalse
To be more precise, in this setting the learning agent
has access to a generative model of the underlying Markov Decision Process (MDP)
and for every state-action pair $(s,a)$,
it can request
a sample of the next state $s'$ that the MDP transitions to.
\fi
This problem 
is similar to understanding the number of labeled examples required in PAC learning \citep{valiant1984theory}.

In this work, we study three different formal
notions of replicability and design algorithms that satisfy them.
First, we study
the definition of \citet{impagliazzo2022reproducibility}, which 
adapted to the context of RL
 says that a learning algorithm is replicable if it outputs
the exact same policy when executed twice on the same MDP, using \emph{shared} internal randomness across the two executions (cf. \Cref{def:replicable policy}). 
We show that there exists a replicable algorithm that outputs 
a near-optimal policy using $\widetilde O(N^3)$ 
samples\footnote{For simplicity, we hide the dependence on the remaining parameters of the problem in this section.},
where $N$ is the cardinality of the state-action space.
This algorithm satisfies an additional property we call \emph{locally random},
which roughly asks that every random decision the algorithm makes
based on internal randomness
must draw its internal randomness independently from other decisions.
Next,
we provide a lower bound for deterministic algorithms
that matches this upper bound.

Subsequently, we study a less stringent notion
of replicability called TV indistinguishability, which was introduced
by \citet{kalavasis2023statistical}. This definition states that, in expectation 
over the random draws of the input, the $\TV$ distance of the two
distributions over the outputs of the algorithm should be small (cf. \Cref{def:tv ind}). We design a computationally efficient TV indistinguishable algorithm
for answering $d$ statistical queries whose sample complexity scales
as $\widetilde O(d^2)$. We remark that this improves the sample complexity
of its replicable counterpart based on the rounding trick 
from \citet{impagliazzo2022reproducibility} by a factor of $d$ and it
has applications outside the scope of our work \citep{impagliazzo2022reproducibility, esfandiari2023replicable,esfandiari2023replicableb,bun2023stability,kalavasis2023statistical}.
This algorithm is inspired by the Gaussian
mechanism from the Differential Privacy (DP) literature \citep{dwork2014algorithmic}. Building upon this statistical
query estimation oracle,
we design computationally efficient 
$\TV$-indistinguishable algorithms for
$Q$-function estimation and policy estimation whose sample complexity
scales as $\widetilde O(N^2).$
Interestingly,
we show that by violating the locally random property
and allowing for internal randomness that creates correlations across decisions,
we can transform these TV indistinguishable algorithms to replicable ones without hurting their sample complexity,
albeit at a cost of $\widetilde O(\exp(N))$ running time. Our transformation
is inspired by the main result of \citet{kalavasis2023statistical}.
We also conjecture that the true sample complexity of $\rho$-replicable policy estimation is indeed $\widetilde \Theta(N^2)$.

Finally,
we propose a novel relaxation of the previous notions of replicability.
Roughly speaking,
we say that an algorithm
is \emph{approximately replicable} if,
with high probability,
when executed twice
on the same MDP,
it outputs policies that are close under a dissimilarity measure
that is based on the \emph{Renyi divergence}.
We remark that this
definition does not require sharing the internal randomness
across the executions.
Finally, we design an RL
algorithm that is approximately replicable and outputs a near-optimal policy
with $\widetilde O(N)$ sample and time complexity.

\begin{table}[h]
    \caption{Complexity Overview for $Q$-Estimation with Constant Probability of Success.}
    \label{tab:Q-estimation complexity overview}
    % \vskip 0.15in
    \begin{center}
    \begin{small}
    \begin{sc}
    \begin{tabular}{lccc}
        \toprule
        Property & Sample Complexity & Time Complexity \\
        \midrule
        Locally Random, Replicable & $\tilde \Theta\left( \frac{N^3}{(1-\gamma)^3 \varepsilon^2 \rho^2} \right)$  & $\tilde \Theta\left( \frac{N^3}{(1-\gamma)^3 \varepsilon^2 \rho^2} \right)$ \\
        TV Indistinguishable & $\tilde O\left( \frac{N^2}{(1-\gamma)^3 \varepsilon^2 \rho^2} \right)$  & $\tilde O\left( \frac{\poly(N)}{(1-\gamma)^3 \varepsilon^2 \rho^2} \right)$ \\
        Replicable (Through TV Indistinguishability) & $\tilde O\left( \frac{N^2}{(1-\gamma)^3 \varepsilon^2 \rho^2} \right)$  & $\tilde O\left( \frac{\exp(N)}{(1-\gamma)^3 \varepsilon^2 \rho^2} \right)$ \\
        \bottomrule
        % \\
        % \toprule
        % Property & Sample Complexity & Time Complexity \\
        % \midrule
        % Locally Random, Replicable & $\tilde O\left( \frac{N^3}{(1-\gamma)^5 \varepsilon^2 \rho^2} \right)$  & $\tilde O\left( \frac{N^3}{(1-\gamma)^5 \varepsilon^2 \rho^2} \right)$ \\
        % TV Indistinguishable & $\tilde O\left( \frac{N^2}{(1-\gamma)^5 \varepsilon^2 \rho^2} \right)$  & $\tilde O\left( \frac{\poly(N)}{(1-\gamma)^5 \varepsilon^2 \rho^2} \right)$ \\
        % Replicable (Through TV Indistinguishability) & $\tilde O\left( \frac{N^2}{(1-\gamma)^5 \varepsilon^2 \rho^2} \right)$  & $\tilde O\left( \frac{\exp(N)}{(1-\gamma)^5 \varepsilon^2 \rho^2} \right)$ \\
        % Approximately Replicable & $\tilde O\left( \frac{N}{(1-\gamma)^5 \varepsilon^2 \rho^2} \right)$ & $\tilde O\left( \frac{N}{(1-\gamma)^5 \varepsilon^2 \rho^2} \right)$ \\
        % \bottomrule
    \end{tabular}
    \end{sc}
    \end{small}
    \end{center}
    % \vskip -0.1in
\end{table}
% \iffalse
\begin{table}[h]
    \caption{Complexity Overview for Policy Estimation with Constant Probability of Success.}
    \label{tab:policy complexity overview}
    % \vskip 0.15in
    \begin{center}
    \begin{small}
    \begin{sc}
    \begin{tabular}{lccc}
        \toprule
        Property & Sample Complexity & Time Complexity \\
        \midrule
        Locally Random, Replicable & $\tilde O\left( \frac{N^3}{(1-\gamma)^5 \varepsilon^2 \rho^2} \right)$  & $\tilde O\left( \frac{N^3}{(1-\gamma)^5 \varepsilon^2 \rho^2} \right)$ \\
        TV Indistinguishable & $\tilde O\left( \frac{N^2}{(1-\gamma)^5 \varepsilon^2 \rho^2} \right)$  & $\tilde O\left( \frac{\poly(N)}{(1-\gamma)^5 \varepsilon^2 \rho^2} \right)$ \\
        Replicable (Through TV Indistinguishability) & $\tilde O\left( \frac{N^2}{(1-\gamma)^5 \varepsilon^2 \rho^2} \right)$  & $\tilde O\left( \frac{\exp(N)}{(1-\gamma)^5 \varepsilon^2 \rho^2} \right)$ \\
        Approximately Replicable & $\tilde O\left( \frac{N}{(1-\gamma)^5 \varepsilon^2 \rho^2} \right)$ & $\tilde O\left( \frac{N}{(1-\gamma)^5 \varepsilon^2 \rho^2} \right)$ \\
        \bottomrule
    \end{tabular}
    \end{sc}
    \end{small}
    \end{center}
    % \vskip -0.1in
\end{table}
% \fi

\Cref{tab:Q-estimation complexity overview} and \Cref{tab:policy complexity overview} 
summarizes the sample and time complexity
of $Q$-estimation and policy estimation,
respectively,
under different notions of replicability.
We assume the algorithms in question have a constant probability of success.
% The sample complexity lower bound only applies to deterministic $\rho$-replicable algorithms
% but our arguments can be applied to locally random $\rho$-replicable $Q$-estimation algorithms.
In \Cref{sec:discussion regarding different repl notions},
we further discuss the benefits and downsides for each of these notions.

\subsection{Related Works}
\textbf{Replicability.}
Pioneered by \citet{impagliazzo2022reproducibility},
there has been a growing interest from the learning theory community in studying
replicability as an algorithmic property. 
\citet{esfandiari2023replicableb, esfandiari2023replicable} studied 
replicable algorithms in the context of multi-armed bandits and clustering.
Recently, \citet{bun2023stability} established equivalences between replicability
and other notions of algorithmic stability such as differential privacy when
the domain of the learning problem is finite and provided some computational
and statistical
hardness results to obtain these equivalences, under cryptographic assumptions. 
Subsequently, \citet{kalavasis2023statistical} proposed a relaxation of the replicability
definition of \citet{impagliazzo2022reproducibility}, 
showed its statistical
equivalence to the notion of replicability for countable domains\footnote{We remark that this equivalence for finite domains can also be obtained, implicitly, from the results of \citet{bun2023stability}.}
and extended some of the equivalences from \citet{bun2023stability} to countable
domains.
\citet{chase2023replicability, dixon2023list} proposed
a notion of \emph{list-replicability},
where the output of the learner is not necessarily identical across two executions
but is limited to a small list of choices.

The closest related work to ours is the concurrent and independent work of \citet{eaton2023replicable}. They also study a formal notion of replicability in RL which is inspired by the work of \citet{impagliazzo2022reproducibility} and coincides with one of the replicability definitions we are studying (cf. \Cref{def:replicability}). Their work focuses both on the generative model and the episodic exploration settings. They derive upper bounds on the sample complexity in both settings and validate their results experimentally. On the other hand, our work focuses solely on the setting with the generative model. We obtain similar sample complexity upper bounds for replicable RL algorithms under \Cref{def:replicability} and then we show a lower bound for the class of locally random algorithms. Subsequently, we consider two relaxed notions of replicability which yield improved sample complexities.

\textbf{Reproducibility in RL.}
Reproducing, interpreting, and evaluating empirical results in RL
can be challenging since there are many sources of randomness in standard benchmark environments.
\citet{khetarpal2018re} proposed a framework for evaluating RL to improve reproducibility.
Another barrier to reproducibility is the unavailability of code and training details within technical reports.
Indeed,
\citet{henderson2018deep} observed that both intrinsic (e.g. random seeds, environments)
and extrinsic (e.g. hyperparameters, codebases) factors can contribute to difficulties in reproducibility.
\citet{tian2019elf} provided an open-source implementation of
AlphaZero \citep{silver2017mastering},
a popular RL-based Go engine. 
% We are not aware of any theoretical works that formally study reproducibility in RL.

\textbf{RL with a Generative Model.}
The study of RL with a generative model was initiated by \citet{kearns1998finite} who provided algorithms with suboptimal sample complexity
in the discount factor $\gamma$. A long line of work (see, e.g. \citet{gheshlaghi2013minimax, wang2017randomized, sidford2018near, sidford2018variance, feng2019does, agarwal2020model,li2020breaking} and
references therein) has led to (non-replicable) algorithms with minimax
optimal sample complexity. 
Another relevant line of work
that culminated with the results of \citet{even2002pac, mannor2004sample}
studied the sample complexity of finding an $\varepsilon$-optimal
arm in the multi-armed bandit setting with access to a generative model.

\section{Setting}
\subsection{Reinforcement Learning Setting}
\textbf{(Discounted) Markov Decision Process.} We start by providing the definitions related to the 
\emph{Markov Decision Process} (MDP) that we study in this work.
\begin{defn}[Discounted Markov Decision Process]
  A \emph{(discounted) Markov decision process (MDP)}
  is a 6-tuple
  $
    M = \left(\mcal S, s_0, \mcal A = \bigcup_{s \in \mcal S}\mcal A^s, P_M, r_M, \gamma\right).
  $
  Here $\mcal S$ is a finite set of states, $s_0 \in \mcal S$ is the
  initial state,
  $\mcal A^s$ is the finite set of available actions for state $s \in \mcal S$,
  and $P_M(s'\mid s, a)$ is the transition kernel, i.e,
  $\forall (s,s') \in \mcal S^2, \forall a \in \mcal A^s, P_M(s'\mid s, a) \geq 0$ and $\forall s \in \mcal S, \forall a \in    \mcal A^s,\sum_{s' \in \mcal S} P_M(s'\mid s, a) = 1$. 
  We denote the reward function\footnote{We assume that the reward is deterministic and known to the learner. Our results hold for stochastic and unknown rewards with an extra (replicable) estimation step, which does not increase the overall sample complexity.} by $r_M: \mcal S \times \mcal A \rightarrow [0,1]$
  and the discount factor by $\gamma\in (0, 1)$. 
  The interaction between
  the agent and the environment works as follows.
  At every step,
  the agent observes a state $s$
  and selects an action $a\in \mcal A^s$,
  yielding an instant reward $r_M(s, a)$.
  The environment then transitions to a random new state $s'\in \mcal S$
  drawn according to the distribution $P_M(\cdot\mid s, a)$.
\end{defn}

\begin{defn}[Policy]
    We say that a map $\pi:\mcal S\to \mcal A$
    is a \emph{(deterministic) stationary policy}.
\end{defn}
When we consider randomized policies we overload the notation and denote
$\pi(s,a)$ the probability mass that policy $\pi$ puts on action $a \in \mcal A^s$ in state $s \in \mcal S.$

\begin{defn}[Value ($V$) Function]
  The \emph{value $(V)$ function} $V_M^\pi: \mcal S\to [0, \nicefrac1{(1-\gamma)}]$
  of a policy $\pi$ with respect to the MDP $M$ is given by
  $
    V_M^\pi(s) := \E\left[ \sum_{t=0}^\infty \gamma^t r_M(s_t, a_t) \mid s_0=s \right].
  $
  Here $a_t \sim \pi(s_t)$ and $s_{t+1}\sim P_M(\cdot\mid s_t, a_t)$.
\end{defn}
This is the expected discounted cumulative reward of a policy.

\begin{defn}[Action-Value ($Q$) Function]
  The \emph{action-value ($Q$) function} $Q_M^\pi: \mcal S \times \mcal A \to [0, \nicefrac1{(1-\gamma)}]$
  of a policy $\pi$ with respect to the MDP $M$ is given by
  $
    Q_M^\pi(s, a) := r_M(s, a) + \gamma \cdot \sum_{s'\in \mcal S} P_M(s'\mid s, a) \cdot  V_M^\pi(s') .
  $
\end{defn}

We write
$N := \sum_{s\in\mcal S} \card{\mcal A^s}$
to denote the number of state-action pairs.
We denote by $\pi^\star$ the \emph{optimal} policy that maximizes the value function,
i.e., $\forall \pi, s\in \mcal S$:
$V^\star(s) := V^{\pi^\star}(s) \geq V^\pi(s)$. We also define $Q^\star(s,a) := Q^{\pi^\star}(s,a).$ This quantity is well defined
since the fundamental theorem of RL states that there exists
a (deterministic) policy $\pi^\star$ that simultaneously maximizes
$V^\pi(s)$ among all policies $\pi$, for all $s \in \mcal S$ (see e.g. \cite{puterman2014markov}). 

Since estimating the optimal policy from samples when $M$ is unknown could be an impossible task,
we aim to compute an $\varepsilon$-\emph{approximately} optimal policy for $M$.
\begin{defn}[Approximately Optimal Policy]\label{def:approximately optimal policy}
    Let $\varepsilon \in (0,1).$ We say that the policy $\pi$ is $\varepsilon$-approximately optimal if
    $||V^\star - V^{\pi}||_{\infty} \leq \varepsilon$.
\end{defn}
In the above definition,
$||\cdot||_\infty$ denotes the infinity norm of the vector, i.e., its maximum element in absolute value.

\textbf{Generative Model.} Throughout this work, we assume we have access to a \emph{generative model} (first studied in \cite{kearns1998finite}) or a \emph{sampler} $G_M$, which takes as input a state-action pair $(s, a)$
and provides a sample $s'\sim P_M(\cdot\mid s, a).$
This widely studied fundamental RL setting allows us to focus on the sample complexity of planning over a long horizon
without considering the additional complications of exploration.
Since our focus throughout this paper is on the \emph{statistical} complexity of the problem,
our goal is to achieve the desired algorithmic performance while 
minimizing the number of samples from the generator that the algorithm
requires.

\textbf{Approximately Optimal Policy Estimator.} We now define 
what it means for an algorithm $\mscr A$ to be an approximately optimal policy estimator. 
\begin{defn}[$(\varepsilon,\delta)$-Optimal Policy Estimator]
    Let $\varepsilon, \delta \in (0,1)^2$. A (randomized) algorithm $\mscr A$ is called
    an $(\varepsilon, \delta)$-optimal policy estimator if there exists a number $n := n(\varepsilon, \delta) \in \N$ such that, for any MDP $M$, when it is given
    at least $n(\varepsilon, \delta)$ samples from the generator $G_M$,
    it outputs a policy
    $\hat\pi$ such that
    $\norm{V^{\hat \pi} - V^\star}_\infty \leq \varepsilon$
    with probability at least $1-\delta$.
    Here, the probability is over random draws from $G_M$ and the internal randomness of $\mscr A$.
\end{defn}
Approximately optimal $V$-function estimators and $Q$-function estimators are defined similarly.

\begin{remark}
    In order to allow flexibility to the algorithm,
    we do not restrict it to request the same amount of samples for every state-action pair. Thus
    $n(\varepsilon,\delta)$ is a bound on the \underline{total} number of samples that 
    $\mscr A$ receives from $G_M.$ The algorithms we design request the same
    number of samples for every state-action pair,
    however, our lower bounds are stronger and hold without this restriction.
\end{remark}

When the MDP $M$ is clear from context,
we omit the subscript in all the previous quantities.

\subsection{Replicability}
\begin{defn}[Replicable Algorithm; \citep{impagliazzo2022reproducibility}]
\label{def:replicability}
  Let $\mscr A: \mcal I^n \rightarrow \mcal O$ be an $n$-sample
  randomized algorithm that takes as input elements from some domain $\mcal I$
  and maps them to some co-domain $\mcal O$.
  Let $\mcal R$ denote the internal distribution over binary strings that $\mscr A$ 
  uses.
  For $\rho \in (0,1)$,
  we say that $\mscr A$ is \emph{$\rho$-replicable} if for any distribution $\mcal D$
  over $\mcal I$ it holds that
  $
    \P_{\bar S, \bar S' \sim \mcal D^n, \bar r \sim \mcal R} \set*{\mscr  A(\bar S; \bar r) = \mscr  A(\bar S'; \bar r)}
    \geq 1-\rho \,,
  $
  where $\mscr A(\bar S; \bar r)$ denotes the (deterministic) output of $\mscr A$ when
  its input is $\bar S$ and the realization of the internal random string is $\bar r$.
\end{defn}
In the context of our work, we should think of  $\mscr A$ as a randomized mapping that receives samples from the generator $G$ and outputs policies.
Thus, even when $\bar S$ is fixed, $\mscr A(\bar S)$ should be thought of as a random variable,
whereas $\mscr A(\bar S; \bar r)$ is the \emph{realization}
of this variable given the (fixed) $\bar S, \bar r$.
We should think of $\bar r$ as the shared randomness between the two executions, which can be implemented as a shared random seed.

One of the most elementary statistical operations
we may wish to make replicable is mean estimation.
This operation can be phrased using the language of \emph{statistical queries}.
\begin{defn}[Statistical Query Oracle; \citep{kearns1998efficient}]\label{def:Statistical Query Oracle} 
    Let $\mathcal{D}$ be a distribution over the domain $\mathcal{X}$
    and $\phi: \mathcal{X}^n \to \R$ be a statistical query
    with true value
    $
        v^\star := \lim_{n\to \infty} \phi(X_1, \dots, X_n)\in \R.
    $
    Here $X_i\sim_{i.i.d.} \mcal D$
    and the convergence is understood in probability or distribution.
    Let $\varepsilon,\delta \in (0,1)^2$.
    A \emph{statistical query (SQ) oracle} outputs 
    a value $v$ such that $\abs*{v - v^\star} \leq \varepsilon$
    with probability at least $1-\delta$.
\end{defn}

The simplest example of a statistical query is the sample mean
$
    \phi(X_1, \dots, X_n)
    = \frac1n \sum_{i=1}^n X_i.
$
\citet{impagliazzo2022reproducibility} designed a replicable SQ-query oracle
for sample mean queries with bounded co-domain (cf. \Cref{thm:replicable mean estimation}).

The following definition is the formal instantiation of \Cref{def:replicability}
in the setting we are studying.
\begin{defn}[Replicable Policy Estimator]\label{def:replicable policy}
  Let $\rho \in (0,1)$. 
  A policy estimator $\mscr A$ that receives samples from a generator $G$
  and returns a policy $\pi$ using internal randomness $\mcal R$
  is $\rho$-replicable
  if for any MDP $M$,
  when two sequences of samples $\bar S, \bar S'$ are generated independently from $G$,
  it holds that
  $
    \P_{\bar S, \bar S' \sim G, \bar r \sim \mcal R} \set*{\mscr A(\bar S; \bar r) = \mscr A (\bar S'; \bar r)}
    \geq 1-\rho.
  $
\end{defn}

To give the reader some intuition about the type of problems for which
replicable algorithms under \Cref{def:replicability} exist,
we consider the fundamental task
of estimating the mean of a random variable.
\citet{impagliazzo2022reproducibility} provided a replicable mean estimation algorithm when the variable is bounded (cf. \Cref{thm:replicable mean estimation}).
\citet{esfandiari2023replicable} generalized the result to simultaneously estimate the means of multiple random variables with unbounded co-domain 
under some regularity conditions on their distributions (cf. \Cref{thm:replicable rounding}).
The idea behind both results is to use a rounding trick introduced in \citet{impagliazzo2022reproducibility} which allows one to sacrifice some accuracy
of the estimator in favor of the replicability property. The formal 
statement of both results,
which are useful for our work,
are deferred to \Cref{apx:omitted replicability preliminaries}.

\section{Replicable \texorpdfstring{$Q$}{Q}-Function \& Policy Estimation}\label{sec:exactly replicable q-function and policy}
Our aim in this section is to understand the sample complexity overhead
that the replicability property imposes on the task of computing an $(\varepsilon,\delta)$-
approximately optimal policy. Without this requirement, 
\citet{sidford2018near,agarwal2020model,li2020breaking} showed that 
\mbox{$\widetilde O( N\log(\nicefrac1\delta) / [(1-\gamma)^3 \varepsilon^2] )$}
samples suffice to estimate such a policy, value function,
and $Q$-function. Moreover, 
since 
\citet{gheshlaghi2013minimax} provided matching lower bounds\footnote{up to logarithmic factors},
the sample complexity for this problem has been settled.
Our main results in this section are tight sample complexity
bounds for locally random $\rho$-replicable $(\varepsilon,\delta)$-approximately optimal $Q$-function estimation
as well as upper and lower bounds for $\rho$-replicable $(\varepsilon,\delta)$-approximately policy
estimation that differ by a factor of $\nicefrac1{(1-\gamma)^2}.$ 
The missing proofs
for this section can be found in \Cref{apx:omitted details approximately replicable section}.

We remark that in both the presented algorithms and lower bounds,
we assume local randomness.
For example,
we assume that the internal randomness is drawn independently 
for each state-action pair for replicable $Q$-estimation.
In the case where we allow for the internal randomness to be correlated across estimated quantities,
we present an algorithm that overcomes our present lower bound in \Cref{sec:from tv ind to repl}.
However, the running time of this algorithm is exponential in $N$.

\subsection{Computationally Efficient Upper Bound on the Sample Complexity}
\label{sec:computationally efficient upper bound and lower bound}
We begin by providing upper bounds on the sample complexity for replicable
estimation of an approximately optimal policy and $Q$-function.
On a high level, we follow a two-step approach:
\begin{enumerate*}[1)]
    \item Start with black-box access to some $Q$-estimation algorithm that is not necessarily replicable (cf. \Cref{thm:variance-reduced QVI})
    to estimate some $\wh{Q}$ such that $\norm*{Q^\star - \wh{Q}}_\infty \leq \varepsilon_0$.
    % and requires $n(\varepsilon,\delta)$ samples to output an $(\varepsilon,\delta)$-approximately optimal $Q$-function,
    \item Apply the replicable rounding algorithm from \Cref{thm:replicable rounding} as a
    post-processing step. 
\end{enumerate*}
The rounding step incurs some loss of accuracy in the estimated $Q$-function.
Therefore, in order to balance between $\rho$-replicability and $(\varepsilon,\delta)$-accuracy,
we need to call the black-box oracle with an accuracy smaller than $\varepsilon$,
i.e. choose $\varepsilon_0 < O(\varepsilon \rho)$.
This yields an increase in the sample complexity
which we quantify below.
% If we use the algorithm from \citet{sidford2018near} as the oracle $\mcal O$, 
% the next result follows immediately.
For the proof details,
see \Cref{apx:replicable Q and policy estimation}.

Recall that $N$ is the number of state-action pairs of the MDP.
\begin{restatable}{thm}{rQEstimation}\label{thm:replicable Q estimation}
    Let $\varepsilon, \rho\in (0, 1)^2$ and $\delta\in (0, \nicefrac\rho3)$.
    There is a locally random $\rho$-replicable algorithm 
    that outputs an $\varepsilon$-optimal $Q$-function with probability at least $1-\delta$.
    Moreover,
    it has time and sample complexity
    $
        \widetilde O( N^3\log(\nicefrac1\delta) / [(1-\gamma)^3 \varepsilon^2 \rho^2] ).
    $
\end{restatable}

So far,
we have provided a replicable algorithm that outputs an approximately optimal
$Q$ function. The main result of \citet{singh1994upper} shows that if 
$\norm*{\wh Q - Q^\star}_\infty \leq \varepsilon$, then the greedy
policy with respect to $\wh Q$, i.e., $\forall s\in \mcal S, \wh \pi(s) := \argmax_{a \in \mcal A^s} \wh Q(s,a)$, is $\nicefrac\varepsilon{(1-\gamma)}$-approximately optimal (cf. \Cref{thm:approx Q to approx policy}). Thus, if we want
to obtain an $\varepsilon$-approximately optimal policy,
it suffices to obtain a
$(1-\gamma)\varepsilon$-approximately optimal $Q$-function. This is formalized
in \Cref{cor:replicable policy sample complexity}.

\begin{restatable}{cor}{rPolicyEstimation}\label{cor:replicable policy sample complexity}
    Let $\varepsilon, \rho\in (0, 1)^2$ and $\delta\in (0, \nicefrac\rho3)$.
    There is a locally random $\rho$-replicable algorithm 
    that outputs an $\varepsilon$-optimal policy with probability at least $1-\delta$.
    Moreover,
    it has time and sample complexity
    $
        \widetilde O( N^3\log(\nicefrac1\delta) / [(1-\gamma)^5 \varepsilon^2 \rho^2] ).
    $
\end{restatable}
Again,
we defer the proof to \Cref{apx:replicable Q and policy estimation}.

Due to space limitation, the lower bound derivation is postponed to \Cref{apx:omitted details approximately replicable section}.

\section{TV Indistinguishable Algorithms for \texorpdfstring{$Q$}{Q}-Function and Policy Estimation}\label{sec:tv ind section}
In this section, we present an algorithm with an improved sample
complexity for replicable $Q$-function estimation and policy estimation. Our approach
consists of several steps. First, we design a computationally efficient 
SQ algorithm for answering
$d$ statistical queries that satisfies
the \emph{total variation
(TV) indistinguishability} property \citep{kalavasis2023statistical} (cf. \Cref{def:tv ind}), which can be viewed as a
relaxation of replicability.
The new SQ algorithm has an improved sample complexity
compared to its replicable counterpart we discussed previously.
Using this oracle, we show how we can design computationally efficient 
$Q$-function estimation and policy estimation
algorithms that satisfy the TV indistinguishability definition
and have an improved sample complexity by a factor of $N$ compared to the ones
in \Cref{sec:computationally efficient upper bound and lower bound}.
Then, by describing a specific implementation 
of its \emph{internal} randomness, we make the algorithm replicable.
Unfortunately, this step incurs an exponential cost in the computational complexity of the
algorithm 
with respect to the cardinality of the state-action space. 
We emphasize that the reason we are able to circumvent the lower bound
of \Cref{sec:lower bound q and policy} is that we use a specific source of internal randomness
that creates correlations across the random choices of the learner. Our result
reaffirms the observation made by \citet{kalavasis2023statistical} that
the same learning algorithm, i.e., input $\to$ output mapping, can be replicable
under one implementation of its internal randomness but not replicable under a different one.

First, we state the definition of TV indistinguishability from \citet{kalavasis2023statistical}.

\begin{defn}[TV Indistinguishability; \citep{kalavasis2023statistical}]\label{def:tv ind}
    A learning rule $\mscr A$ is $n$-sample $\rho$-$\TV$ indistinguishable if for any distribution over inputs $\mcal D$ and two independent samples $S, S' \sim \mcal D^n$ it holds that
$
\E_{S,S' \sim \mcal D^n} [ \dtv( A(S), A(S') ) ] \leq \rho\,.
$
\end{defn}

In their work, \citet{kalavasis2023statistical} showed how to transform
any $\rho$-$\TV$ indistinguishable algorithm to a $2\rho/(1+\rho)$-replicable
one when the input
domain is \emph{countable}. Importantly, this transformation does not change 
the input $\to$ output mapping that is induced by the algorithm.
A similar transformation for finite domains can also be obtained by the results
in \citet{bun2023stability}. We emphasize that neither of these two transformations
are computationally efficient. Moreover, \citet{bun2023stability} give cryptographic evidence
that there might be an inherent computational hardness to obtain the transformation.

\subsection{TV Indistinguishable Estimation of Multiple Statistical Queries}
\label{sec:tv ind estimation of multiple SQs}
We are now ready to present a $\TV$-indistinguishable algorithm for estimating
$d$ independent statistical queries. The high-level approach is as follows. First,
we estimate each statistical query up to accuracy $\nicefrac{\varepsilon \rho}{\sqrt{d}}$
using black-box access to
the SQ oracle and we get an estimate $\widehat{\mu}_1 \in [0,1]^d$. Then, the output
of the algorithm is drawn from $\mcal N(\widehat{\mu}_1, \varepsilon^2 I_d).$ Since the 
estimated mean of each query is accurate up to $\nicefrac{\varepsilon \rho}{\sqrt{d}}$ and the variance
is $\varepsilon^2$, we can see that, with high probability, the estimate of 
each query will be accurate up to $O(\varepsilon).$ To argue about the
TV indistinguishability property, we first notice that, with high probability across
the two executions, the estimate $\widehat{\mu}_2 \in [0,1]^d$ satisfies
$||\widehat{\mu}_1 - \widehat{\mu}_2||_\infty \leq 2\rho\cdot\varepsilon/\sqrt{d}.$
Then, we can bound the TV distance of the output of the algorithm as
$\dtv\left(\mcal N(\widehat{\mu}_1, \varepsilon^2 I_d),
\mcal N(\widehat{\mu}_2, \varepsilon^2 I_d)\right) \leq O(\rho)$ \citep{gupta2020kl}. 
We underline
that this behavior is reminiscent of the advanced composition theorem in the Differential Privacy (DP)
literature (see e.g., \citet{dwork2014algorithmic}) and our algorithm can be viewed
as an extension of the Gaussian mechanism from the DP line of work to the replicability setting.
This algorithm has applications outside the scope of our work since multiple statistical query
estimation is a subroutine widely used in the replicability line of work \citep{impagliazzo2022reproducibility, esfandiari2023replicable, esfandiari2023replicableb, bun2023stability, kalavasis2023statistical}.
This discussion is formalized in the 
following theorem.
\begin{restatable}[TV Indistinguishable SQ Oracle for Multiple Queries]{thm}{tvMultipleSQ}\label{thm:tv ind
oracle for multiple queries}
    Let $\varepsilon, \rho \in (0,1)^2$ and $\delta \in (0, \nicefrac\rho5).$ Let $\phi_1,\ldots,\phi_d$
    be $d$ statistical queries with co-domain $[0, 1]$.
    Assume that we can simultaneously estimate the true values of all $\phi_i$'s 
    with accuracy $\varepsilon$
    and confidence $\delta$ using $n(\varepsilon, \delta)$ total samples. 
    Then, there exists a
    $\rho$-$\TV$ indistinguishable algorithm (\Cref{alg:tv ind for multiple queries}) that requires at most
    $
        n( \nicefrac{\varepsilon \rho}{[2\sqrt{8d\cdot\log(4d/\delta)}]}, \nicefrac\delta{2} )
    $
    many samples to output estimates $\widehat{v}_1, \ldots, \widehat{v}_d$ 
    of the true values $v_1, \dots, v_d$ to guarantee that 
    $
        \max_{i \in [d]} \abs{\widehat{v}_i - v_i} \leq \varepsilon \,,
    $
    with probability at least $1-\delta$.
\end{restatable}

\subsection{TV Indistinguishable \texorpdfstring{$Q$}{Q}-Function and Policy Estimation}
\label{sec:tv ind q estimation and policy estimation}
Equipped with \Cref{alg:tv ind for multiple queries}, we are now 
ready to present a $\TV$-indistinguishable algorithm for $Q$-function estimation
and policy estimation with superior sample complexity compared to the one
in \Cref{sec:computationally efficient upper bound and lower bound}. The idea is similar to the one in \Cref{sec:computationally efficient upper bound and lower bound}. We start with black-box access to an algorithm for $Q$-function estimation, and
then we apply the Gaussian mechanism (\Cref{alg:tv ind for multiple queries}).
We remark that the running time of this algorithm is polynomial in all the parameters of the
problem.

Recall that $N$ is the number of state-action pairs of the MDP.
\begin{restatable}{thm}{tvIndQEstimation}\label{thm:tv ind Q estimation}
    Let $\varepsilon, \rho\in (0, 1)^2$ and $\delta\in (0, \nicefrac\rho5)$.
    There is a $\rho$-$\TV$ indistinguishable algorithm 
    that outputs an $\varepsilon$-optimal $Q$-function with probability at least $1-\delta$.
    Moreover,
    it has time and sample complexity
    $
        \widetilde O( N^2\log(\nicefrac1\delta) / [(1-\gamma)^3 \varepsilon^2 \rho^2] ).
    $
\end{restatable}

\begin{pf}
    The proof follows by combining the guarantees of \citet{sidford2018near} (\Cref{thm:variance-reduced QVI}) and \Cref{thm:tv ind
oracle for multiple queries}. To be more precise, \Cref{thm:variance-reduced QVI} shows
that in order to compute some $\widehat{Q}$ such that
$
    \norm*{\widehat{Q} - Q}_{\infty} \leq \varepsilon \,,
$
one needs
$
    \widetilde O( N\log(\nicefrac1\delta) / [(1-\gamma)^3 \varepsilon^2] ).
$
Thus, in order
to apply \Cref{thm:tv ind
oracle for multiple queries} the 
sample complexity becomes
$
    \widetilde O( N^2\log(\nicefrac1\delta) / [(1-\gamma)^3 \varepsilon^2 \rho^2] ).
$
\end{pf}

Next, we describe a TV indistinguishable algorithm that enjoys similar sample complexity
guarantees. Similarly as before, we use the main result of \citet{singh1994upper} 
which shows that if 
$\norm*{\wh Q - Q^\star}_\infty \leq \varepsilon$, then the greedy
policy with respect to $\wh Q$, i.e., $\forall s\in \mcal S, \wh \pi(s) := \argmax_{a \in \mcal A^s} \wh Q(s,a)$, is $\nicefrac\varepsilon{(1-\gamma)}$-approximately optimal (cf. \Cref{thm:approx Q to approx policy}). Thus, if we want
to obtain an $\varepsilon$-approximately optimal policy,
it suffices to obtain a
$(1-\gamma)\varepsilon$-approximately optimal $Q$-function. 
The indistinguishable guarantee follows from the data-processing inequality
This is formalized
in \Cref{cor:tv ind policy sample complexity}.

\begin{restatable}{cor}{tvIndPolicyEstimation}\label{cor:tv ind policy sample complexity}
    Let $\varepsilon, \rho\in (0, 1)^2$ and $\delta\in (0, \nicefrac\rho5)$.
    There is a $\rho$-$TV$ indistinguishable algorithm 
    that outputs an $\varepsilon$-optimal policy with probability at least $1-\delta$.
    Moreover,
    it has time and sample complexity
    $
        \widetilde O( N^2\log(\nicefrac1\delta) / [(1-\gamma)^5 \varepsilon^2 \rho^2] ).
    $
\end{restatable}

\subsection{From TV Indistinguishability to Replicability}
\label{sec:from tv ind to repl}
We now describe how we can transform the TV indistinguishable algorithms we provided
to replicable ones. As we alluded to before, 
this transformation does not hurt the sample complexity, but
requires exponential time in the state-action space. 
Our transformation is based on the approach proposed by \citet{kalavasis2023statistical}
which holds when the input domain is \emph{countable}. 
Its main idea is that when two random variables follow distributions that are $\rho$-close
in $\TV$-distance, then there is a way to couple them using only \emph{shared randomness}.
The implementation of this coupling is based on the \emph{Poisson point process} and can
be thought of a generalization of von Neumann's rejection-based sampling
to handle more general domains.
We underline that in general spaces without structure 
it is not known yet how to obtain such a coupling. However, even though the input domain
of the Gaussian mechanism is \emph{uncountable} and the result of \citet{kalavasis2023statistical} does not apply directly in our setting, we are able
to obtain a similar transformation as they did. The main step required to perform
this transformation is to find a reference measure with respect to which
the algorithm is \emph{absolutely continuous}. We provide these crucial measure-theoretic 
definitions below.

\begin{defn}[Absolute Continuity]
  Consider two measures $P, \mcal P$ on a $\sigma$-algebra $\mcal B$ of subsets of $W$.
  We say that $P$ is absolutely continuous with respect to $\mcal P$ 
  if for any $E \in \mcal B$ such that $\mcal P(E) = 0$, it holds that $P(E) = 0$. 
\end{defn}

Recall that $\mscr A(S)$ denotes the \emph{distribution} over outputs, when the input to the algorithm is $S.$

\begin{defn}\label{def:absolutely cts learning rule for any D}
    Given a learning rule $\mscr A$ and reference probability measure $\mcal P$, we say that $\mscr A$ is absolutely continuous with respect to $\mcal P$ if for any input $S$,  
    $\mscr A(S)$ is absolutely continuous with respect to $\mcal P$.
\end{defn}
We emphasize that this property should hold for every fixed sample $S$,
i.e.,
the randomness of the samples are not taken into account.

We now define what it means for two learning rules to be \emph{equivalent}.
\begin{defn}[Equivalent Learning Rules]\label{def:equivalent 
learning rules}
    Two learning rules $\mscr A, \mscr A'$ are \emph{equivalent} if for every fixed sample $S$,
    it holds that $\mscr A(S) \eq{d} \mscr A'(S)$, 
    i.e., for the same input they induce the same distribution over outputs.
\end{defn}

Using a coupling technique based on the Poisson point process,
we can convert the TV indistinguishable learning algorithms we have proposed so far 
to equivalent ones that are replicable.
See \Cref{alg:pairwise optimal coupling} for a description of how to output a sample from this coupling.
Let us view $\mscr A(S; r), \mscr A(S'; r)$ as random vectors
with small TV distance.
The idea is to implement the shared internal randomness $r$ using rejection sampling
so that the ``accepted'' sample will be the same across two executions
with high probability.
\iffalse
Note that this is non-trivial since the rejection sampling in $\mscr A(S; r)$
can be a function of the data $S$,
which most likely differs from the data $S'$
in another execution.
\fi
For some background regarding the Poisson
point process and the technical tools we use, we refer the reader to \Cref{sec:coupling}.

Importantly, 
for every $S$, 
the output $\mscr A(S)$ of the algorithms we have proposed in \Cref{sec:tv ind estimation of multiple SQs}
and \Cref{sec:tv ind q estimation and policy estimation}
follow a Gaussian distribution, 
which is absolutely continuous with respect to the Lebesgue measure. 
Furthermore, 
the Lebesgue measure is $\sigma$-finite so we can use the coupling algorithm (cf. \Cref{alg:pairwise optimal coupling})
of \citet{angel2019pairwise},
whose guarantees are stated in \Cref{thm:pairwise opt coupling protocol}.
We are now ready to state the result regarding the improved $\rho$-replicable
SQ oracle for multiple queries. Its proof is an adaptation of the main
result of \citet{kalavasis2023statistical}.

\begin{restatable}[Replicable SQ Oracle for Multiple Queries]{thm}{rSQByPPP}\label{thm:improved replicable
oracle for multiple queries}
    Let $\varepsilon, \rho \in (0,1)^2$ and $\delta \in (0, \nicefrac\rho5).$ Let $\phi_1,\ldots,\phi_d$
    be $d$ statistical queries with co-domain $[0, 1]$.
    Assume that we can simultaneously estimate the true values of all $\phi_i$'s 
    with accuracy $\varepsilon$
    and confidence $\delta$ using $n(\varepsilon, \delta)$ total samples. 
    Then, there exists a
    $\rho$-replicable algorithm that requires at most
    $
        n( \nicefrac{\varepsilon \rho}{[4\sqrt{8d\cdot\log(4d/\delta)}]}, \nicefrac\delta{2} )
    $
    many samples to output estimates $\widehat{v}_1, \ldots, \widehat{v}_d$ 
    of the true values $v_1, \dots, v_d$ 
    with the guarantee that 
    $
        \max_{i \in [d]} \abs{\widehat{v}_i - v_i} \leq \varepsilon \,,
    $
    with probability at least $1-\delta$.
\end{restatable}

By using an identical argument, we can obtain $\rho$-replicable algorithms
for $Q$-function estimation and policy estimation.
Recall that $N$ is the number of state-action pairs of the MDP.
\begin{restatable}{thm}{ImprovedRQEstimation}\label{thm:improved replicable Q estimation}
    Let $\varepsilon, \rho\in (0, 1)^2$ and $\delta\in (0, \nicefrac\rho4)$.
    There is a $\rho$-replicable algorithm 
    that outputs an $\varepsilon$-optimal $Q$-function with probability at least $1-\delta$.
    Moreover,
    it has sample complexity
    $
        \widetilde O( N^2\log(\nicefrac1\delta) / [(1-\gamma)^3 \varepsilon^2 \rho^2] ).
    $
\end{restatable}

\begin{restatable}{cor}{improvedRPolicyEstimation}\label{cor:improved replicable policy sample complexity}
    Let $\varepsilon, \rho\in (0, 1)^2$ and $\delta\in (0, \nicefrac\rho4)$.
    There is a $\rho$-replicable algorithm 
    that outputs an $\varepsilon$-optimal policy with probability at least $1-\delta$.
    Moreover,
    it has sample complexity
    $
        \widetilde O( N^2\log(\nicefrac1\delta) / [(1-\gamma)^5 \varepsilon^2 \rho^2] ).
    $
\end{restatable}

In \Cref{rem:coordinate wise coupling 
of Gaussian mechanism} we explain why we cannot
use a coordinate-wise coupling.

\section{Approximately Replicable Policy Estimation}\label{sec:approximately replicable policy estimation}
    The definitions of replicability (cf. \Cref{def:replicable policy}, \Cref{def:tv ind}) 
we have discussed
so far suffer from a significant sample complexity blow-up in terms of the cardinality of the state-action space 
which can be prohibitive in many settings of interest. 
% Another potential issue with this definition is the need to store explicitly the random seed 
% that was used in the execution of the algorithm 
% and share it across multiple different runs. 
In this section, 
we propose \emph{approximate replicability},
a relaxation of these definitions, 
and show that this property can be achieved
with a significantly milder sample
complexity compared to (exact) replicability.
Moreover,
this definition does not require shared internal randomness across the executions of the algorithm.

% First, it is instructive to state a stronger definition
% of algorithmic stability that was introduced by \citet{bun2020equivalence}.
% \begin{defn}[Global Stability; \citep{bun2020equivalence}]\label{def:global stability}
%     Let $\mcal X$ be some input domain. We say that a learning algorithm $\mscr A$ is
%     $\rho$-globally stable if for any distribution $\mcal D$ 
%     over $\mcal X$ it holds that
%       \[
%         \P_{S,S' \sim \mcal D^n, r, r' \sim \mcal R}\set{\mscr A(S;r) = \mscr A(S';r')} \geq 
%         1 - \rho \,.
%     \]
% \end{defn}
% We underline that, unlike \Cref{def:replicability}, the internal randomness across
% two executions of the algorithm is \emph{not} shared. Interestingly, \cite{chase2023replicability}
% show that for various tasks the stability parameter of an algorithm
% that satisfies \Cref{def:global stability}
% cannot be boosted, whereas, as shown in~\citet{impagliazzo2022reproducibility},
% the replicability parameter can always be boosted.

First, we define a general notion of \emph{approximate} replicability as follows.
\begin{defn}[Approximate Replicability]\label{def:approximate replicability}
    Let $\mcal X, \mcal Y$ be the input and output domains, respectively.
    Let $ \kappa: \mcal Y \times \mcal Y \rightarrow \R_{\geq 0}$ be some distance function on $\mcal Y$ and let $\rho_1, \rho_2 \in (0,1)^2$. 
    We say that an algorithm $\mscr A$
    is $(\rho_1, \rho_2)$-approximately replicable with respect to $\kappa$ if for any distribution $\mcal D$ over $\mcal X$ it holds that
    $
        \P_{S, S' \sim \mcal D^n, r, r' \sim \mcal R}\set{\kappa(\mscr A(S;r), \mscr A(S';r')) \geq \rho_1} \leq \rho_2 \,. 
    $
\end{defn}
In words, this relaxed version of \Cref{def:replicability} requires that 
the outputs of the algorithm, when executed on two sets of i.i.d. data, 
using \emph{independent} internal randomness across the two executions,
are close under some appropriate distance measure. 
In the context of our work,
the output of the learning algorithm is some 
policy $\pi: \mcal S \rightarrow \Delta(\mcal A)$,
where $\Delta(\mcal A)$ denotes
the probability simplex over $\mcal A$.
Thus,
it is natural to instantiate $\kappa$ as some \emph{dissimilarity measure} of distributions
like the total variation (TV) distance
or the Renyi divergence.
For the exact definition of these dissimilarity measures,
we refer the reader to \Cref{apx:omitted-defs}.
We now state the definition of an approximately replicable 
policy estimator.

\begin{defn}[Approximately Replicable Policy Estimator]\label{def:approx-replicable-policy}
    Let $\mscr A$ be an algorithm 
    that takes as input samples
    of state-action pair transitions and returns a policy $\pi$. 
    Let $\kappa$ be some dissimilarity measure on $\Delta(\mcal A)$
    and let $\rho_1,\rho_2 \in (0,1)^2$. 
    We say that $\mscr A$ is $(\rho_1,\rho_2)$-approximately replicable
    if for any MDP $M$ it holds that
    $
        \P_{S,S'\sim G, r,r' \sim \mcal R}\left\{\max_{s\in \mcal S}
        \kappa(\pi(s), \pi'(s)) \geq \rho_1 \right\} \leq \rho_2 \,,
    $
    where $G$ is the generator of state-action pair transitions, $\mcal R$ is the source of internal randomness
    of $\mscr A$,
    $\pi$ is the output of $\mscr A$ on input $S, r$, and $\pi'$ is its
    output on input $S', r'.$
\end{defn}

To the best of our knowledge,
the RL algorithms that have been developed for the model we are studying do not satisfy this property. 
Nevertheless, many of them compute
an estimate $Q$ with the promise that $ \norm{Q-Q^\star}_\infty \leq \varepsilon$ \citep{sidford2018near, agarwal2020model, li2020breaking}. Thus, it is not 
hard to see that if we run the algorithm twice on independent
data with independent internal randomness we have that
$\norm{Q-Q'}_\infty \leq 2\varepsilon$.
This is exactly the main property that we need
in order to obtain approximately replicable policy estimators. The key idea
is that instead
of outputting the greedy policy with respect to this $Q$-function,
we output a policy given by some \emph{soft-max} rule. Such a rule
is a mapping 
$\R^{\mcal A}_{\geq 0} \rightarrow \Delta(\mcal A)$
that achieves two desiderata:
\begin{enumerate*}[(i)]
    \item The distribution over the actions is ``stable'' with 
    respect to perturbations of the $Q$-function.
    \item For every $s \in \mcal S$,
    the value of the policy $V^\pi(s)$
    that is induced by this mapping is ``close'' to $V^\star(s)$.
\end{enumerate*}

Formally, the stability of the
soft-max rule is captured through its Lipschitz constant
(cf. \Cref{def:lipschitz-cont}).
In this setting,
this means that whenever the two functions $Q, Q'$ are close
under some distance measure (e.g. the $\ell_\infty$ norm),
then the policies that are induced by the soft-max rule are close
under some (potentially different) dissimilarity measure. 
The approximation guarantees of the soft-max rules
are captured by the following definition.
\begin{defn}[Soft-Max Approximation; \citep{epasto2020optimal}]\label{def:soft-max approx}
    Let $\varepsilon > 0.$ A soft-max function $f: \R^{\mcal A}_{\geq 0} \rightarrow \Delta(\mcal A)$ is \emph{$\varepsilon$-approximate}
    if for all $x \in \R^{\mcal A}$,
    $\iprod{f(x), x} \geq \max_{a \in \mcal A} x_a - \varepsilon$.
    % Similarly, we say that $f$ is \emph{worst case $\varepsilon$-approximate} if 
    % \[
    %     \forall x \in \R^{|\mcal A|}:\quad f(x_a) > 0 \implies x_a \geq \max_{a' \in \mcal A} x_{a'} - \varepsilon \,.
    % \]
\end{defn}

In this work,
we focus on the soft-max rule that is induced by the exponential function (\ExpSoftMax),
which has been studied in
several application domains \citep{gibbs1902elementary, mcsherry2007mechanism, huang2012exponential, dwork2014algorithmic, gao2017properties}.
Recall $\pi(s,a)$ denotes the probability mass that policy $\pi$ puts on action $a \in \mcal A^s$ in state $s \in \mcal S$.
Given some $\lambda > 0$ and $Q(s,a) \in \R^{\mcal S \times \mcal A}$,
the induced randomized policy $\pi$ is given by
$
     \pi(s,a) = \frac{\exp{\lambda Q(s,a)}}{\sum_{a' \in \mcal A^s} \exp{\lambda Q(s,a')}} \,.
$
For a discussion about the advantages of using
more complicated soft-max rules like the one developed in \citet{epasto2020optimal},
we refer the reader to \Cref{apx:different
soft-max rules}. 
% It turns
% out that when we require replicability with respect to the Renyi divergence,
% \ExpSoftMax achieves the optimal $\varepsilon$-approximation
% guarantees and there are no soft-max rules that are worst case $\varepsilon$-
% approximately optimal.

We now describe our results
when we consider approximate replicability with respect to the Renyi 
divergence and the Total Variation (TV) distance. At a high level, 
our approach is divided into two steps:
\begin{enumerate*}[1)]
    \item Run some $Q$-learning algorithm (e.g. \citep{sidford2018near, agarwal2020model, li2020breaking})
    to estimate some $\wh{Q}$ such that $\norm*{Q^\star - \wh{Q}}_\infty \leq \varepsilon$.
    \item Estimate the policy using some soft-max rule. 
\end{enumerate*}
One advantage of this approach is that it allows for flexibility
and different implementations of these steps that better suit the application
domain. An important lemma we use is the 
following.
% \subsection{Approximate Replicability Under Renyi Divergence}
% \label{sec:approx-replicability-renyi}
% We are now ready to explain how we obtain approximately replicable 
% policy estimators under the Renyi divergence. We use the \ExpSoftMax rule
% to derive this result
% (cf. \Cref{eq:exp-soft-max}). First, we state the approximation guarantee
% we get from this rule.

\begin{lem}[Exponential Soft-Max Approximation Guarantee; \citep{mcsherry2007mechanism}]\label{lem:exp-soft-max-approx}
    Let $\varepsilon \in (0,1), \alpha, p \geq 1$, and set
    $\lambda = \nicefrac{\log(d)}{\varepsilon}$, where $d$ is the ambient
    dimension of the input domain. Then, \ExpSoftMax with parameter $\lambda$
    is $\varepsilon$-approximate and $2\lambda$-Lipschitz continuous (cf. \Cref{def:lipschitz-cont})
    with respect to $(\ell_p, D_\alpha)$, where $D_\alpha$ is
    the Renyi divergence of order $\alpha.$
\end{lem}
This is an important building block of our proof.
However, 
it is not sufficient on its own in order to bound the gap of the \ExpSoftMax
policy and the optimal one. This is handled in the next lemma whose
proof is postponed to \Cref{apx:approximately replicable policy}. Essentially,
it can be viewed as an extension of the result in \citet{singh1994upper}
to handle the soft-max policy instead of the greedy one.
\begin{restatable}[Soft-Max Policy vs Optimal Policy]{lem}{arSoftmaxPolicy}\label{lem:exp-soft-max-policy}
    Let $\varepsilon_1, \varepsilon_2 \in (0,1)^2$. Let $\wh{Q} \in \R^{\mcal S\times \mcal A}$ be such that 
    $\norm*{\wh Q - Q^\star} \leq \varepsilon_1$.
    Let $\hat\pi$ be the \ExpSoftMax policy with
    respect to $\wh Q$ using parameter $\lambda = \nicefrac{\log\card{\mcal A}}{\varepsilon_2}$.
    Then,
    $\norm{V^{\hat \pi} - V^\star}_{\infty}
        \leq \nicefrac{(2\varepsilon_1 + \varepsilon_2)}{(1-\gamma)}$.
\end{restatable}

\iffalse
We present the main result in this section.
We show that there exists an approximately replicable algorithm 
(cf. \Cref{alg:approximately replicable policy estimation}) that outputs an 
approximately optimal policy and enjoys a mild blow-up 
in its sample complexity compared to its non-replicable counterparts.
The algorithm works in two steps:
First,
it estimates an approximately optimal $\wh Q$-function.
Then,
it outputs a policy given by the soft-max function
with respect to $\wh Q$.
\fi

Combining \Cref{lem:exp-soft-max-approx} and \Cref{lem:exp-soft-max-policy} yields the desired approximate replicability guarantees we seek.
The formal proof of the following result is postponed to \Cref{apx:approximately replicable policy}. 
Recall we write $N := \sum_{s\in \mcal S} \card{\mcal A^s}$
to denote the total number of state-action pairs.
\begin{restatable}{thm}{arPolicyEstimation}\label{thm:approximately-replicable-policy-estimation-sample
-complexity}
    Let $\alpha \geq 1, \gamma, \delta, \rho_1, \rho_2 \in (0,1)^4$,
    and $\varepsilon \in \left(0,(1-\gamma)^{-1/2}\right)$. There is a $(\rho_1, \rho_2)$-approximately
    replicable algorithm $\mscr A$ with respect to the Renyi divergence $D_\alpha$
    such that given access to a generator $G$ for any MDP $M$,
    it outputs a policy $\hat \pi$
    for which $\norm*{V^{\hat \pi} - V^\star}_\infty \leq \varepsilon$
    with probability at least $1-\delta$. Moreover,
    $\mscr A$ has time and sample complexity
    $
        \widetilde O(N\log(\nicefrac{1}{\min\{\delta,\rho_2\}}) / [(1-\gamma)^5 \varepsilon^2 \rho_1^2]) \,.
    $
\end{restatable}

% The following remarks regarding this result are worth mentioning.

\section{Conclusion}\label{sec:conclusion}
In this work,
we  establish sample complexity bounds for a several notions of replicability
in the context of RL. We give an extensive comparison  of the guarantees under these different notions
in \Cref{sec:discussion regarding different repl notions}. We
believe that our work can open several directions for future research.
One immediate next step would be to verify our lower bound conjecture
for replicable estimation of multiple independent coins (cf. \Cref{conj:replicable randomized coin lower bound}). Moreover, it would be very
interesting to extend our results to different RL settings,
e.g. offline RL with linear MDPs, offline RL with finite horizon, and 
online RL.

%\section*{Acknowledgements}
\begin{ack}
    Amin Karbasi acknowledges funding in direct support of this work from NSF (IIS-1845032), ONR (N00014-19-1-2406), and the AI Institute for Learning-Enabled Optimization at Scale (TILOS). Lin Yang is supported in part by NSF Award 2221871 and an Amazon Faculty Award. 
    Grigoris Velegkas is supported by TILOS, the Onassis Foundation, and the Bodossaki Foundation. Felix Zhou is supported by TILOS. The authors would also like to thank Yuval Dagan for an insightful discussion regarding
the Gaussian mechanism.
\end{ack}

%\clearpage
\bibliography{references}

\begin{thebibliography}{52}
\providecommand{\natexlab}[1]{#1}
\providecommand{\url}[1]{\texttt{#1}}
\expandafter\ifx\csname urlstyle\endcsname\relax
  \providecommand{\doi}[1]{doi: #1}\else
  \providecommand{\doi}{doi: \begingroup \urlstyle{rm}\Url}\fi

\bibitem[Afsar et~al.(2022)Afsar, Crump, and Far]{afsar2022reinforcement}
M~Mehdi Afsar, Trafford Crump, and Behrouz Far.
\newblock Reinforcement learning based recommender systems: A survey.
\newblock \emph{ACM Computing Surveys}, 55\penalty0 (7):\penalty0 1--38, 2022.

\bibitem[Agarwal et~al.(2020)Agarwal, Kakade, and Yang]{agarwal2020model}
Alekh Agarwal, Sham Kakade, and Lin~F Yang.
\newblock Model-based reinforcement learning with a generative model is minimax
  optimal.
\newblock In \emph{Conference on Learning Theory}, pages 67--83. PMLR, 2020.

\bibitem[Angel and Spinka(2019)]{angel2019pairwise}
Omer Angel and Yinon Spinka.
\newblock Pairwise optimal coupling of multiple random variables.
\newblock \emph{arXiv preprint arXiv:1903.00632}, 2019.

\bibitem[Baker(2016)]{baker20161}
Monya Baker.
\newblock 1,500 scientists lift the lid on reproducibility.
\newblock \emph{Nature}, 533\penalty0 (7604), 2016.

\bibitem[Bavarian et~al.(2016)Bavarian, Ghazi, Haramaty, Kamath, Rivest, and
  Sudan]{bavarian2016optimality}
Mohammad Bavarian, Badih Ghazi, Elad Haramaty, Pritish Kamath, Ronald~L Rivest,
  and Madhu Sudan.
\newblock Optimality of correlated sampling strategies.
\newblock \emph{arXiv preprint arXiv:1612.01041}, 2016.

\bibitem[Bertsekas(1976)]{bertsekas1976dynamic}
Dimitri~P Bertsekas.
\newblock \emph{Dynamic programming and stochastic control.}
\newblock Academic Press, 1976.

\bibitem[Broder(1997)]{broder1997resemblance}
Andrei~Z Broder.
\newblock On the resemblance and containment of documents.
\newblock In \emph{Proceedings. Compression and Complexity of SEQUENCES 1997
  (Cat. No. 97TB100171)}, pages 21--29. IEEE, 1997.

\bibitem[Bun et~al.(2023)Bun, Gaboardi, Hopkins, Impagliazzo, Lei, Pitassi,
  Sorrell, and Sivakumar]{bun2023stability}
Mark Bun, Marco Gaboardi, Max Hopkins, Russell Impagliazzo, Rex Lei, Toniann
  Pitassi, Jessica Sorrell, and Satchit Sivakumar.
\newblock Stability is stable: Connections between replicability, privacy, and
  adaptive generalization.
\newblock \emph{arXiv preprint arXiv:2303.12921}, 2023.

\bibitem[Charikar(2002)]{charikar2002similarity}
Moses~S Charikar.
\newblock Similarity estimation techniques from rounding algorithms.
\newblock In \emph{Proceedings of the thiry-fourth annual ACM symposium on
  Theory of computing}, pages 380--388, 2002.

\bibitem[Chase et~al.(2023)Chase, Moran, and
  Yehudayoff]{chase2023replicability}
Zachary Chase, Shay Moran, and Amir Yehudayoff.
\newblock Replicability and stability in learning.
\newblock \emph{arXiv preprint arXiv:2304.03757}, 2023.

\bibitem[Dixon et~al.(2023)Dixon, Pavan, Woude, and
  Vinodchandran]{dixon2023list}
Peter Dixon, A~Pavan, Jason~Vander Woude, and NV~Vinodchandran.
\newblock List and certificate complexities in replicable learning.
\newblock \emph{arXiv preprint arXiv:2304.02240}, 2023.

\bibitem[Dwork et~al.(2014)Dwork, Roth, et~al.]{dwork2014algorithmic}
Cynthia Dwork, Aaron Roth, et~al.
\newblock The algorithmic foundations of differential privacy.
\newblock \emph{Foundations and Trends{\textregistered} in Theoretical Computer
  Science}, 9\penalty0 (3--4):\penalty0 211--407, 2014.

\bibitem[Eaton et~al.(2023)Eaton, Hussing, Kearns, and
  Sorrell]{eaton2023replicable}
Eric Eaton, Marcel Hussing, Michael Kearns, and Jessica Sorrell.
\newblock Replicable reinforcement learning.
\newblock \emph{arXiv preprint arXiv:2305.15284}, 2023.

\bibitem[Epasto et~al.(2020)Epasto, Mahdian, Mirrokni, and
  Zampetakis]{epasto2020optimal}
Alessandro Epasto, Mohammad Mahdian, Vahab Mirrokni, and Emmanouil Zampetakis.
\newblock Optimal approximation-smoothness tradeoffs for soft-max functions.
\newblock \emph{Advances in Neural Information Processing Systems},
  33:\penalty0 2651--2660, 2020.

\bibitem[Esfandiari et~al.(2023{\natexlab{a}})Esfandiari, Kalavasis, Karbasi,
  Krause, Mirrokni, and Velegkas]{esfandiari2023replicableb}
Hossein Esfandiari, Alkis Kalavasis, Amin Karbasi, Andreas Krause, Vahab
  Mirrokni, and Grigoris Velegkas.
\newblock Replicable bandits.
\newblock In \emph{The Eleventh International Conference on Learning
  Representations}, 2023{\natexlab{a}}.

\bibitem[Esfandiari et~al.(2023{\natexlab{b}})Esfandiari, Karbasi, Mirrokni,
  Velegkas, and Zhou]{esfandiari2023replicable}
Hossein Esfandiari, Amin Karbasi, Vahab Mirrokni, Grigoris Velegkas, and Felix
  Zhou.
\newblock Replicable clustering.
\newblock \emph{arXiv preprint arXiv:2302.10359}, 2023{\natexlab{b}}.

\bibitem[Even-Dar et~al.(2002)Even-Dar, Mannor, and Mansour]{even2002pac}
Eyal Even-Dar, Shie Mannor, and Yishay Mansour.
\newblock Pac bounds for multi-armed bandit and markov decision processes.
\newblock In \emph{COLT}, volume~2, pages 255--270. Springer, 2002.

\bibitem[(FAIR)† et~al.(2022)(FAIR)†, Bakhtin, Brown, Dinan, Farina,
  Flaherty, Fried, Goff, Gray, Hu, et~al.]{meta2022human}
Meta Fundamental AI Research Diplomacy~Team (FAIR)†, Anton Bakhtin, Noam
  Brown, Emily Dinan, Gabriele Farina, Colin Flaherty, Daniel Fried, Andrew
  Goff, Jonathan Gray, Hengyuan Hu, et~al.
\newblock Human-level play in the game of diplomacy by combining language
  models with strategic reasoning.
\newblock \emph{Science}, 378\penalty0 (6624):\penalty0 1067--1074, 2022.

\bibitem[Feng et~al.(2019)Feng, Yin, and Yang]{feng2019does}
Fei Feng, Wotao Yin, and Lin~F Yang.
\newblock How does an approximate model help in reinforcement learning?
\newblock \emph{arXiv preprint arXiv:1912.02986}, 2019.

\bibitem[Gao and Pavel(2017)]{gao2017properties}
Bolin Gao and Lacra Pavel.
\newblock On the properties of the softmax function with application in game
  theory and reinforcement learning.
\newblock \emph{arXiv preprint arXiv:1704.00805}, 2017.

\bibitem[Gheshlaghi~Azar et~al.(2013)Gheshlaghi~Azar, Munos, and
  Kappen]{gheshlaghi2013minimax}
Mohammad Gheshlaghi~Azar, R{\'e}mi Munos, and Hilbert~J Kappen.
\newblock Minimax pac bounds on the sample complexity of reinforcement learning
  with a generative model.
\newblock \emph{Machine learning}, 91:\penalty0 325--349, 2013.

\bibitem[Gibbs(1902)]{gibbs1902elementary}
Josiah~Willard Gibbs.
\newblock \emph{Elementary principles in statistical mechanics: developed with
  especial reference to the rational foundations of thermodynamics}.
\newblock C. Scribner's sons, 1902.

\bibitem[Gupta(2020)]{gupta2020kl}
Rishabh Gupta.
\newblock Kl divergence between 2 gaussian distributions, 2020.
\newblock URL
  \url{https://mr-easy.github.io/2020-04-16-kl-divergence-between-2-gaussian-distributions/}.

\bibitem[Henderson et~al.(2018)Henderson, Islam, Bachman, Pineau, Precup, and
  Meger]{henderson2018deep}
Peter Henderson, Riashat Islam, Philip Bachman, Joelle Pineau, Doina Precup,
  and David Meger.
\newblock Deep reinforcement learning that matters.
\newblock In \emph{Proceedings of the AAAI conference on artificial
  intelligence}, volume~32, 2018.

\bibitem[Holenstein(2007)]{holenstein2007parallel}
Thomas Holenstein.
\newblock Parallel repetition: simplifications and the no-signaling case.
\newblock In \emph{Proceedings of the thirty-ninth annual ACM symposium on
  Theory of computing}, pages 411--419, 2007.

\bibitem[Huang and Kannan(2012)]{huang2012exponential}
Zhiyi Huang and Sampath Kannan.
\newblock The exponential mechanism for social welfare: Private, truthful, and
  nearly optimal.
\newblock In \emph{2012 IEEE 53rd Annual Symposium on Foundations of Computer
  Science}, pages 140--149. IEEE, 2012.

\bibitem[Impagliazzo et~al.(2022)Impagliazzo, Lei, Pitassi, and
  Sorrell]{impagliazzo2022reproducibility}
Russell Impagliazzo, Rex Lei, Toniann Pitassi, and Jessica Sorrell.
\newblock Reproducibility in learning.
\newblock \emph{arXiv preprint arXiv:2201.08430}, 2022.

\bibitem[Kalavasis et~al.(2023)Kalavasis, Karbasi, Moran, and
  Velegkas]{kalavasis2023statistical}
Alkis Kalavasis, Amin Karbasi, Shay Moran, and Grigoris Velegkas.
\newblock Statistical indistinguishability of learning algorithms.
\newblock \emph{arXiv preprint arXiv:2305.14311}, 2023.

\bibitem[Kearns(1998)]{kearns1998efficient}
Michael Kearns.
\newblock Efficient noise-tolerant learning from statistical queries.
\newblock \emph{Journal of the ACM (JACM)}, 45\penalty0 (6):\penalty0
  983--1006, 1998.

\bibitem[Kearns and Singh(1998)]{kearns1998finite}
Michael Kearns and Satinder Singh.
\newblock Finite-sample convergence rates for q-learning and indirect
  algorithms.
\newblock \emph{Advances in neural information processing systems}, 11, 1998.

\bibitem[Khetarpal et~al.(2018)Khetarpal, Ahmed, Cianflone, Islam, and
  Pineau]{khetarpal2018re}
Khimya Khetarpal, Zafarali Ahmed, Andre Cianflone, Riashat Islam, and Joelle
  Pineau.
\newblock Re-evaluate: Reproducibility in evaluating reinforcement learning
  algorithms.(2018).
\newblock In \emph{International conference on machine learning}. ICML, 2018.

\bibitem[Kiran et~al.(2021)Kiran, Sobh, Talpaert, Mannion, Al~Sallab, Yogamani,
  and P{\'e}rez]{kiran2021deep}
B~Ravi Kiran, Ibrahim Sobh, Victor Talpaert, Patrick Mannion, Ahmad~A
  Al~Sallab, Senthil Yogamani, and Patrick P{\'e}rez.
\newblock Deep reinforcement learning for autonomous driving: A survey.
\newblock \emph{IEEE Transactions on Intelligent Transportation Systems},
  23\penalty0 (6):\penalty0 4909--4926, 2021.

\bibitem[Kleinberg and Tardos(2002)]{kleinberg2002approximation}
Jon Kleinberg and Eva Tardos.
\newblock Approximation algorithms for classification problems with pairwise
  relationships: Metric labeling and markov random fields.
\newblock \emph{Journal of the ACM (JACM)}, 49\penalty0 (5):\penalty0 616--639,
  2002.

\bibitem[Levin and Peres(2017)]{levin2017markov}
David~A Levin and Yuval Peres.
\newblock \emph{Markov chains and mixing times}, volume 107.
\newblock American Mathematical Soc., 2017.

\bibitem[Li et~al.(2020)Li, Wei, Chi, Gu, and Chen]{li2020breaking}
Gen Li, Yuting Wei, Yuejie Chi, Yuantao Gu, and Yuxin Chen.
\newblock Breaking the sample size barrier in model-based reinforcement
  learning with a generative model.
\newblock \emph{Advances in neural information processing systems},
  33:\penalty0 12861--12872, 2020.

\bibitem[Mannor and Tsitsiklis(2004)]{mannor2004sample}
Shie Mannor and John~N Tsitsiklis.
\newblock The sample complexity of exploration in the multi-armed bandit
  problem.
\newblock \emph{Journal of Machine Learning Research}, 5\penalty0
  (Jun):\penalty0 623--648, 2004.

\bibitem[McSherry and Talwar(2007)]{mcsherry2007mechanism}
Frank McSherry and Kunal Talwar.
\newblock Mechanism design via differential privacy.
\newblock In \emph{48th Annual IEEE Symposium on Foundations of Computer
  Science (FOCS'07)}, pages 94--103. IEEE, 2007.

\bibitem[Mnih et~al.(2013)Mnih, Kavukcuoglu, Silver, Graves, Antonoglou,
  Wierstra, and Riedmiller]{mnih2013playing}
Volodymyr Mnih, Koray Kavukcuoglu, David Silver, Alex Graves, Ioannis
  Antonoglou, Daan Wierstra, and Martin Riedmiller.
\newblock Playing atari with deep reinforcement learning.
\newblock \emph{arXiv preprint arXiv:1312.5602}, 2013.

\bibitem[Ouyang et~al.(2022)Ouyang, Wu, Jiang, Almeida, Wainwright, Mishkin,
  Zhang, Agarwal, Slama, Ray, et~al.]{ouyang2022training}
Long Ouyang, Jeffrey Wu, Xu~Jiang, Diogo Almeida, Carroll Wainwright, Pamela
  Mishkin, Chong Zhang, Sandhini Agarwal, Katarina Slama, Alex Ray, et~al.
\newblock Training language models to follow instructions with human feedback.
\newblock \emph{Advances in Neural Information Processing Systems},
  35:\penalty0 27730--27744, 2022.

\bibitem[Pineau et~al.(2019)Pineau, Sinha, Fried, Ke, and
  Larochelle]{pineau2019iclr}
Joelle Pineau, Koustuv Sinha, Genevieve Fried, Rosemary~Nan Ke, and Hugo
  Larochelle.
\newblock Iclr reproducibility challenge 2019.
\newblock \emph{ReScience C}, 5\penalty0 (2):\penalty0 5, 2019.

\bibitem[Pineau et~al.(2021)Pineau, Vincent-Lamarre, Sinha, Larivi{\`e}re,
  Beygelzimer, d’Alch{\'e} Buc, Fox, and Larochelle]{pineau2021improving}
Joelle Pineau, Philippe Vincent-Lamarre, Koustuv Sinha, Vincent Larivi{\`e}re,
  Alina Beygelzimer, Florence d’Alch{\'e} Buc, Emily Fox, and Hugo
  Larochelle.
\newblock Improving reproducibility in machine learning research: a report from
  the neurips 2019 reproducibility program.
\newblock \emph{Journal of Machine Learning Research}, 22, 2021.

\bibitem[Puterman(2014)]{puterman2014markov}
Martin~L Puterman.
\newblock \emph{Markov decision processes: discrete stochastic dynamic
  programming}.
\newblock John Wiley \& Sons, 2014.

\bibitem[Sidford et~al.(2018{\natexlab{a}})Sidford, Wang, Wu, Yang, and
  Ye]{sidford2018near}
Aaron Sidford, Mengdi Wang, Xian Wu, Lin Yang, and Yinyu Ye.
\newblock Near-optimal time and sample complexities for solving markov decision
  processes with a generative model.
\newblock \emph{Advances in Neural Information Processing Systems}, 31,
  2018{\natexlab{a}}.

\bibitem[Sidford et~al.(2018{\natexlab{b}})Sidford, Wang, Wu, and
  Ye]{sidford2018variance}
Aaron Sidford, Mengdi Wang, Xian Wu, and Yinyu Ye.
\newblock Variance reduced value iteration and faster algorithms for solving
  markov decision processes.
\newblock In \emph{Proceedings of the Twenty-Ninth Annual ACM-SIAM Symposium on
  Discrete Algorithms}, pages 770--787. SIAM, 2018{\natexlab{b}}.

\bibitem[Silver et~al.(2017)Silver, Schrittwieser, Simonyan, Antonoglou, Huang,
  Guez, Hubert, Baker, Lai, Bolton, et~al.]{silver2017mastering}
David Silver, Julian Schrittwieser, Karen Simonyan, Ioannis Antonoglou, Aja
  Huang, Arthur Guez, Thomas Hubert, Lucas Baker, Matthew Lai, Adrian Bolton,
  et~al.
\newblock Mastering the game of go without human knowledge.
\newblock \emph{nature}, 550\penalty0 (7676):\penalty0 354--359, 2017.

\bibitem[Singh and Yee(1994)]{singh1994upper}
Satinder~P Singh and Richard~C Yee.
\newblock An upper bound on the loss from approximate optimal-value functions.
\newblock \emph{Machine Learning}, 16:\penalty0 227--233, 1994.

\bibitem[Tian et~al.(2019)Tian, Ma, Gong, Sengupta, Chen, Pinkerton, and
  Zitnick]{tian2019elf}
Yuandong Tian, Jerry Ma, Qucheng Gong, Shubho Sengupta, Zhuoyuan Chen, James
  Pinkerton, and Larry Zitnick.
\newblock Elf opengo: An analysis and open reimplementation of alphazero.
\newblock In \emph{International conference on machine learning}, pages
  6244--6253. PMLR, 2019.

\bibitem[Valiant(1984)]{valiant1984theory}
Leslie~G Valiant.
\newblock A theory of the learnable.
\newblock \emph{Communications of the ACM}, 27\penalty0 (11):\penalty0
  1134--1142, 1984.

\bibitem[Vinyals et~al.(2019)Vinyals, Babuschkin, Czarnecki, Mathieu, Dudzik,
  Chung, Choi, Powell, Ewalds, Georgiev, et~al.]{vinyals2019grandmaster}
Oriol Vinyals, Igor Babuschkin, Wojciech~M Czarnecki, Micha{\"e}l Mathieu,
  Andrew Dudzik, Junyoung Chung, David~H Choi, Richard Powell, Timo Ewalds,
  Petko Georgiev, et~al.
\newblock Grandmaster level in starcraft ii using multi-agent reinforcement
  learning.
\newblock \emph{Nature}, 575\penalty0 (7782):\penalty0 350--354, 2019.

\bibitem[Wang(2017)]{wang2017randomized}
Mengdi Wang.
\newblock Randomized linear programming solves the discounted markov decision
  problem in nearly-linear (sometimes sublinear) running time.
\newblock \emph{arXiv preprint arXiv:1704.01869}, 2017.

\bibitem[Yao(1977)]{yao1977probabilistic}
Andrew Chi-Chin Yao.
\newblock Probabilistic computations: Toward a unified measure of complexity.
\newblock In \emph{18th Annual Symposium on Foundations of Computer Science
  (sfcs 1977)}, pages 222--227. IEEE Computer Society, 1977.

\bibitem[Yu et~al.(2021)Yu, Liu, Nemati, and Yin]{yu2021reinforcement}
Chao Yu, Jiming Liu, Shamim Nemati, and Guosheng Yin.
\newblock Reinforcement learning in healthcare: A survey.
\newblock \emph{ACM Computing Surveys (CSUR)}, 55\penalty0 (1):\penalty0 1--36,
  2021.

\end{thebibliography}
\bibliographystyle{plainnat}

\clearpage
\appendix
\iffalse
\section{Useful Facts}
\begin{prop}[Bretagnolle-Huber-Carol Inequality; \citet{vaart1997weak}]\label{prop:BHC inequality}
  Suppose the random vector $(Z^{(1)}, \dots, Z^{(N)})$ is multinomially distributed
  with parameters $(p^{(1)}, \dots, p^{(N)})$ and $n$.
  Let $\wh p^{(j)} := \frac1n Z^{(j)}$.
  Then
  \[
    \P\set*{\sum_{j=1}^N \abs{\wh p^{(j)} - p^{(j)}} \geq 2\varepsilon}
    \leq 2^N \exp\left( -2\varepsilon^2 n \right).
  \]

  In particular,
  for any $\varepsilon\in (0, 1)$,
  sampling
  \[
    n \geq \frac{\ln\frac1\rho + N\ln 2}{2\varepsilon^2}
  \]
  points from a finite distribution
  implies that $\sum_{j=1}^N \abs{\wh p^{(j)} - p^{(j)}} < 2\varepsilon$
  with probability at least $1-\rho$.
\end{prop}
\fi

\section{Omitted Algorithms}\label{apx:omitted-algos}
\begin{algorithm}[h]
\caption{TV Indistinguishable Oracle for Multiple Query Estimation}\label{alg:tv ind for multiple queries}
\begin{algorithmic}[1]
   \STATE $\widehat{\mu} = \left( \widehat{\mu_1},\ldots,\widehat{\mu_d} \right) \gets \mathrm{StatisticalQueryOracles}\left( \frac{\varepsilon\rho}{2\sqrt{8d\cdot\log(4d/\delta)}}, \frac\delta{2} \right)$
   \STATE Sample $\widehat{v} \sim \mcal N(\widehat{\mu}, \varepsilon^2/(8\cdot \log(4d/\delta))\cdot I_d)$
   \STATE Output $\wh v$
\end{algorithmic}
\end{algorithm}

\begin{algorithm}[H]
\caption{Sampling from Pairwise Optimal Coupling; \citep{angel2019pairwise}}\label{alg:pairwise optimal coupling}
\begin{algorithmic}[1]
  \STATE {\bfseries Input:}{ collection of random vectors $\mcal S = \set{X}$ absolutely continuous with respect to a $\sigma$-finite measure $\mu$, with densities $f_X: \R^d\to \R$, some $X\in \mcal S$}
  \STATE Let $\mcal R$ denote the Poisson point process over $\R^d\times \R_+\times \R_+$
  with intensity $\mu\times \Leb\times \Leb$.
  \STATE Sample $r := \set{(x_i, y_i, t_i): i\in \N} \sim \mcal R$.
  \STATE Let $i^\star \gets \argmin_{i\in \N} \set{t_i: f_S(x_i) > y_i}$.
  \STATE Output $x_{i^\star}$ as a sample for $X$.
\end{algorithmic}
\end{algorithm}
\section{Omitted Definitions}\label{apx:omitted-defs}
In this section,
we discuss some dissimilarity measures for probability distributions
that we use in this work.

\begin{defn}[Total Variation (TV) Distance]\label{def:tv-distance}
  Let $\Omega$ be a countable domain and $P, Q$ be probability 
  distributions over $\Omega$. The total variation distance
  between $P, Q$,
  denoted by $d_{\text{TV}}(P,Q)$ is defined as
  \begin{align*}
      d_{\text{TV}}(P,Q) &= \sup_{A \in 2^\Omega} P(A) - Q(A) \\
                         &= ||P-Q||_1 \\
                         &= \inf_{(X,Y) \sim \Pi(P,Q)} \P \set{X \neq Y} \,,
  \end{align*}  
  where $\Pi(P,Q)$ is a \emph{coupling} between $P,Q.$
\end{defn}
In the above definition, a coupling is defined to be a joint probability distribution $(X, Y)\sim \Pi(P, Q)$
over the product space whose marginals distributions are $P, Q$,
i.e., if we only view the individual components,
then $X \sim P, Y \sim Q.$

\begin{defn}[Renyi Divergence]\label{def:renyi-divergence}
   Let $\Omega$ be a countable domain and $P, Q$ be probability 
  distributions over $\Omega$. For any $\alpha > 1$ the Renyi divergence
  of order $\alpha$ between $P, Q$,
  denoted by $D_\alpha(P\|Q)$ is defined as 
  \[
    D_a(P\|Q) = \frac{1}{\alpha - 1}\log \left(\sum_{\omega \in \Omega} \frac{P(\omega)^\alpha}{Q(\omega)^{\alpha-1   }} \right) \,.
  \]
\end{defn}

We note that the quantity above is undefined when $\alpha = 1.$ However,
we can take its limit and define $D_1(P||Q) = \sum_{\omega \in \Omega} P(\omega)
\log\frac{P(\omega)}{Q(\omega)}$, which recovers the definition of KL-divergence.
Similarly, we can define
$D_\infty(P||Q) = \max_{\omega \in \Omega}\frac{P(\omega)}{Q(\omega)}.$

Another important definition is that of Lipschitz continuity.

\begin{defn}[Lipschitz Continuity]\label{def:lipschitz-cont}
    Let $\mcal X, \mcal Y$ be two domains,
    $f:\mcal X \rightarrow \mcal Y$ be some function,
    and $d_1: \mcal X \times \mcal X \rightarrow \R_{\geq 0}$,
    $d_2: \mcal Y \times \mcal Y \rightarrow \R_{\geq 0}$ be distance measures over $\mcal X, \mcal Y$, 
    respectively. We say that $f$ is $L$-Lipschitz continuous
    with respect to $d_1, d_2$ if $\forall x_1, x_2 \in \mcal X$
    it holds that \mbox{$d_2(f(x_1),f(x_2)) \leq L \cdot d_1(x_1,x_2)$}.
\end{defn}

\subsection{Local Randomness}
Our algorithms in \Cref{sec:exactly replicable q-function and policy} satisfy a property which we call \emph{locally random}.
This roughly means that for every decision an algorithm makes based on external and internal randomness,
the internal randomness is used once and discarded immediately after.
\begin{defn}[Locally Random]\label{def:locally random}
  Let $\mscr A = (\mscr A^{(1)}, \dots, \mscr A^{(N)}): \mcal I^n\to \R^N$ be an $n$-sample
  randomized algorithm that takes as input elements from some domain $\mcal I$
  and maps them to $\R^N$.
  We say that $\mscr A$ is \emph{locally random} if:
  \begin{enumerate}[(i)]
      \item The $i$-th output component $\mscr A^{(i)}(\bar S; \bar r^{(i)})$ is a function of all samples $\bar S$ but only its own internal random string $\bar r^{(i)}$.
      \item The sources $\bar r^{(i)}$ of internal randomness are independent of each other
      and the external samples $\bar S$.
  \end{enumerate}
\end{defn}

We will see that by restricting ourselves to locally random algorithms,
it is necessary and sufficent to incur a sample cost of $\tilde \Theta(N^3)$
for replicable $Q$-estimation.
However,
by relaxing this restriction
and allowing for internal randomness that is correlated,
we can achieve $\tilde O(N^2)$ sample complexity.

\section{Omitted Preliminaries}\label{apx:omitted preliminaries}
\subsection{Replicability}\label{apx:omitted replicability preliminaries}
\citet{impagliazzo2022reproducibility} introduced the definition of replicability
and demonstrated the following basic replicable operation.
% We first need to define the \emph{statistical query} model \citep{kearns1998efficient}. 
% \lin{I feel the following formal definition does not help explain the existence of the algorithm in 2.8. Maybe make it more intuitive. Do we need this formal definition later on?}

\begin{thm}[Replicable SQ-Oracle; \citep{impagliazzo2022reproducibility}]\label{thm:replicable mean estimation}
  Let $\varepsilon, \rho\in (0, 1)$ and $\delta\in (0, \nicefrac\rho3)$.
  Suppose $\phi$ is a sample mean statistical query with co-domain $[0, 1]$.
  There is a locally random $\rho$-replicable SQ oracle to estimate its true value
  with tolerance $\varepsilon$ and failure rate $\delta$.
  Moreover,
  the oracle has sample complexity
  \[
    \tilde O\left( \frac1{\varepsilon^2 \rho^2} \log\frac1\delta \right).
  \]
\end{thm}

\citet{esfandiari2023replicable} generalized the result above to multiple general queries with unbounded co-domain,
assuming some regularity conditions on the queries.
\begin{thm}[Replicable Rounding; \citep{impagliazzo2022reproducibility, esfandiari2023replicable}]\label{thm:replicable rounding}
  Let $\varepsilon, \rho\in (0, 1)$ and $\delta\in (0, \nicefrac\rho3)$.
  Suppose we have a finite class of statistical queries $g_1, \dots, g_N$
  with true values $G_1, \dots, G_N$
  and sampling $n$ independent points from $\mcal D$ ensures that
  \[
    \sum_{j=1}^N \abs{g_j(x_1, \dots, x_n) - G_j} \leq \varepsilon' := \frac{\varepsilon(\rho - 2\delta)}{\rho+1-2\delta}
  \]
  with probability at least $1-\delta$.

  There is a locally random $\rho$-replicable algorithm
  that outputs estimates $\wh G_j$
  such that
  \[
    \abs{\wh G_j - G_j}\leq \varepsilon
  \]
  with probability at least $1-\delta$ for every $j\in [N]$.
  Moreover,
  it requires at most $n$ samples.
\end{thm}

\subsection{Coupling and Correlated Sampling} 
\label{sec:coupling}
Our exposition in this section follows \citet{kalavasis2023statistical}.
Coupling is a fundamental notion in probability theory with many applications \citep{levin2017markov}. The correlated sampling problem, which has applications in various domains, e.g., in sketching and approximation algorithms \citep{broder1997resemblance,charikar2002similarity}, is described in \citet{bavarian2016optimality} as follows: Alice and Bob are assigned probability distributions $P_A$ and $P_B$, respectively, over a finite set $W$. \emph{Without any communication, using only shared
randomness} as the means to coordinate, 
Alice is required to output an element $x$ distributed according to $P_A$ and Bob is required to output an element $y$ distributed according to $P_B$. 
Their goal is to minimize the disagreement probability
$\P\set{x \neq y}$, which is comparable with $\dtv(P_A, P_B)$. 
Formally, a correlated sampling strategy for a finite set $W$ 
with error $\varepsilon : [0,1] \to [0,1]$ is specified by a probability space $\mcal R$ 
and a pair of functions $f,g : \Delta_W \times \mcal R \to W$, 
which are measurable in their second argument, such that for any pair $P_A, P_B \in \Delta_W$ with $\dtv(P_A, P_B) \leq \delta$,
it holds that 
\begin{enumerate*}[(i)]
    \item the push-forward measure $f(P_A, \cdot)$ (resp. $g(P_B, \cdot)$) is $P_A$ (resp. $P_B$) and
    \item $\P_{r \sim \mcal R} \set{f(P_A, r) \neq g(P_B, r)} \leq \varepsilon(\delta)$.
\end{enumerate*}
We underline that a correlated sampling strategy is \emph{not} the same as a coupling,
in the sense that the latter requires a single function $h : \Delta_W \times \Delta_W \to \Delta_{W \times W}$ such that for any $P_A, P_B$, the marginals of $h(P_A, P_B)$ are $P_A$ and $P_B$ respectively.

It is known that for any coupling function $h$, it holds 
that $\P_{(x,y) \sim h(P_A, P_B)}\set{x \neq y} \geq \dtv(P_A, P_B)$ 
and that this bound is attainable. Since $(f(P_A, \cdot), g(P_B, \cdot))$
induces a coupling, it holds that $\varepsilon(\delta) \geq \delta$
and, perhaps surprisingly, there exists a strategy with $\varepsilon(\delta) \leq \frac{2\delta}{1+\delta}$ \citep{broder1997resemblance,kleinberg2002approximation,holenstein2007parallel} and this result is tight \citep{bavarian2016optimality}.
A second difference between coupling and correlated sampling has to do with the size of $W$: while correlated sampling strategies can be extended to infinite spaces $W$, it remains open whether there exists a correlated sampling strategy
for general measure spaces $(W, \mcal F, \mu)$ with any non-trivial error bound \citep{bavarian2016optimality}.
On the other hand, coupling applies to spaces $W$ of any size. For further comparisons between coupling and the correlated sampling problem of \cite{bavarian2016optimality}, we refer to the discussion in \cite{angel2019pairwise} after Corollary 4.

\begin{defn}
[Coupling]
\label{def:coupling}
A coupling of two probability distributions $P_A$ and
$P_B$ is a pair of random variables $(X, Y)$, defined on the same probability space, such that the marginal distribution of $X$ is $P_A$ and the marginal distribution of $Y$ is $P_B$.
\end{defn}

A very useful tool for our derivations is a coupling protocol that can be found
in \cite{angel2019pairwise}.

\begin{thm}[Pairwise Optimal Coupling \citep{angel2019pairwise}]\label{thm:pairwise opt coupling protocol}
    Let $\mcal S$ be any collection of random variables that are absolutely continuous with
    respect to a $\sigma$-finite measure $\mu$. 
    Then, there exists a coupling of the
    variables in $\mcal S$
    such that for any $X, Y \in \mcal S$,
    \[
        \P\set{X \neq Y} \leq \frac{2\dtv(X,Y)}{1+ \dtv(X,Y)} \,.
    \]
    Moreover, this coupling requires sample access to a Poisson point
    process with intensity $\mu \times \mathrm{Leb} \times \mathrm{Leb}$, where
    $\mathrm{Leb}$ is the Lebesgue measure over $\R_+,$ and full access
    to the densities of all the random variables in $\mcal S$ with respect to $\mu$.
    Finally,
    we can sample from this coupling using \Cref{alg:pairwise optimal coupling}.
\end{thm}

\section{Omitted Details from \texorpdfstring{\Cref{sec:exactly replicable q-function and policy}}{Replicable Q-Function \& Policy Estimation}}\label{apx:omitted details approximately replicable section}
Here we fill in the details from \Cref{sec:exactly replicable q-function and policy},
which describes upper and lower bounds for replicable $Q$-function and policy estimation.
In \Cref{apx:replicable Q and policy estimation},
we provide the rigorous analysis of an algorithm for locally random replicable $Q$-function estimation
as well as replicable policy estimation.
\Cref{apx:coin problem} describes the locall random replicable version of the multiple coin estimation problem,
an elementary statistical problem
that serves as the basis of our hardness proofs.
Next,
\Cref{apx:replicable Q lower bound} reduces locally random replicable $Q$-function estimation to the multiple coin estimation problem.
Finally,
\Cref{apx:replicable policy lower bound} reduces deterministic replicable policy estimation to locally random replicable $Q$-function estimation.

\subsection{Omitted Details from Upper Bounds}\label{apx:replicable Q and policy estimation}
First,
we show that we can use any non-replicable $Q$-function estimation algorithm as a black-box
to obtain a locally random replicable $Q$-function estimation algorithm. 

\begin{lem}\label{lem:replicable Q rounding}
    Let $\varepsilon', \delta'\in (0, 1)$.
    Suppose there is an oracle $\mscr O$ that takes
    $\widetilde O(f(N, \varepsilon', \delta'))$
    samples from the generative model
    and outputs $\wh Q$ satisfying
    $\norm*{\wh Q - Q^\star}_\infty \leq \varepsilon'$
    with probability at least $1-\delta'$.

    Let $\varepsilon, \rho\in (0, 1)$ and $\delta\in (0, \nicefrac\rho3)$.
    There is a  locally random $\rho$-replicable algorithm which makes a single call to $\mscr O$ and outputs some $\bar Q$ satisfying
    $\norm*{\bar Q - Q^\star}_\infty \leq \varepsilon$
    with probability at least $1-\delta$.
    Moreover,
    it has sample complexity
    $\widetilde O\left( f \left(N, \nicefrac{\varepsilon_0}{N}, \delta \right) \right)$
    where
    $\varepsilon_0 := \nicefrac{\varepsilon(\rho - 2\delta)}{(\rho+1-2\delta)}$.
\end{lem}

\begin{pf}[\Cref{lem:replicable Q rounding}]
    Consider calling $\mscr O$ with $\widetilde O(f(N, \nicefrac{\varepsilon_0}N, \delta))$ samples.
    This ensures that
    \[
        \sum_{s, a} \abs*{\wh Q(s, a) - Q^\star(s, a)} \leq \varepsilon_0
    \]
    with probability at least $1-\delta$.
    By \Cref{thm:replicable rounding},
    there is a locally random $\rho$-replicable algorithm that makes a single call to $\mscr O$ with the number of samples above
    and outputs $\bar Q(s, a)$ such that
    \[
      \norm{\bar Q - Q^\star}_\infty \leq \varepsilon
    \]
    with a probability of success of at least $1-\delta$.
\end{pf}

\citet{sidford2018near} showed a (non-replicable) $Q$ function estimation algorithm
which has optimal (non-replicable) sample complexity
up to logarithmic factors.  

Recall we write $N := \sum_{s\in \mcal S} \card{\mcal A^s}$
to denote the total number of state-action pairs.
\begin{thm}[\citep{sidford2018near}]\label{thm:variance-reduced QVI}
    Let $\varepsilon, \delta\in (0, 1)$,
    there is an algorithm that outputs an $\varepsilon$-optimal policy,
    $\varepsilon$-optimal value function,
    and $\varepsilon$-optimal $Q$-function
    with probability at least $1-\delta$
    for any MDP.
    Moreover,
    it has time and sample complexity
    \[
        \widetilde O\left( \frac{N}{(1-\gamma)^3 \varepsilon^2}\log \frac1\delta \right).
    \]
\end{thm}

The proof of \Cref{thm:replicable Q estimation},
which we repeat below for convenience,
follows by combining \Cref{lem:replicable Q rounding}
and \Cref{thm:variance-reduced QVI}.
\rQEstimation*

The following result of \citet{singh1994upper} relates the $Q$-function estimation error to the quality of the greedy policy
with respect to the estimated $Q$-function.
\begin{thm}[\citep{singh1994upper}]\label{thm:approx Q to approx policy}
    Let $\wh Q$ be a $Q$-function such that $||\wh Q - Q^\star||_\infty \leq \varepsilon.$ Let $\pi(s) := \arg\max_{a \in \mcal S}\wh Q(s,a), \forall s \in \mcal S.$ Then,
    \[
        \norm{V^{\pi} - V^\star}_\infty \leq \frac{\varepsilon}{1-\gamma} \,.
    \]
\end{thm}

\Cref{thm:approx Q to approx policy} enables us to prove \Cref{cor:replicable policy sample complexity},
an upper bound on the sample complexity of replicably estimating an $\varepsilon$-optimal policy.
We restate the corollary below for convenience.
\rPolicyEstimation*

\begin{pf}[\Cref{cor:replicable policy sample complexity}]
    Apply the $\rho$-replicable algorithm from \Cref{thm:replicable Q estimation}
    to yield an $\varepsilon_0$-optimal $Q$-function
    and output the greedy policy based on this function.
    \Cref{thm:approx Q to approx policy} guarantees that the greedy policy derived from an $\varepsilon_0$-optimal $Q$-function is $\nicefrac{\varepsilon_0}{(1-\gamma)}$-optimal.
    Choosing $\varepsilon_0 := (1-\gamma)\varepsilon$ yields the desired result.

    The replicability follows from the fact that conditioned on the event that
    the two $Q$-functions across the two runs are the same, which happens with 
    probability at least $1-\rho$,
    the greedy policies with respect to 
    the underlying $Q$-functions will also be the same\footnote{assuming a consistent tie-breaking rule}.
\end{pf}

\subsection{Lower Bounds for Replicable \texorpdfstring{$Q$}{Q}-Function \& Policy Estimation}\label{sec:lower bound q and policy}
We now move on to the lower bounds
and our approaches to obtain them. 
First, we describe a sample complexity lower bound for locally random $\rho$-replicable algorithms
that seek to estimate $Q^\star$.
Then,
we reduce policy estimation to $Q$-estimation.
Since the dependence of the sample complexity
on the confidence parameter $\delta$ 
of the upper bound is at most polylogarithmic, 
the main focus of the lower bound is on the dependence on the size of 
the state-action space $N$,
the error parameter $\varepsilon$,
the replicability parameter $\rho$,
and the discount factor $\gamma$.

\subsubsection{Intuition of the \texorpdfstring{$Q$}{Q}-Function Lower Bound}
Our MDP construction that witnesses the lower bound relies
on the sample complexity lower bound 
for locally random algorithms that replicably estimate the biases of \emph{multiple independent} coins. 
\citet{impagliazzo2022reproducibility} showed that any $\rho$-replicable
algorithm that estimates the bias of a \emph{single} coin with
accuracy $\varepsilon$ requires at least $\Omega(\nicefrac1{\rho^2\varepsilon^2})$
samples (cf. \Cref{thm:SQ-lowerbound}). We generalize this result
and derive a lower bound for any locally random $\rho$-replicable algorithm that estimates
the biases of $N$ coins with accuracy $\varepsilon$
and constant probability of success.
\iffalse
Ignoring the replicability constraint,
generalizing the sample complexity lower bound of estimating the bias of a single coin
to a lower bound for multiple independent coins is not too difficult.
This is because the union bound is asymptotically tight for independent events.
However,
reasoning about two executions with shared randomness requires a more subtle treatment.
\fi
We discuss our approach in \Cref{sec:coin problem}.
% Now,
% samples from each of the $N$ transition probabilities are independent.
% Intuitively speaking,
% estimating the $Q$-function replicably should be at least as hard as replicably estimating the bias of $N$ independent coins.

Next, given some $\varepsilon, \rho, \gamma, N$, 
we design an MDP for which estimating an approximately optimal
$Q$-function is at least as hard as estimating $N$ coins.
The main technical challenge for this part of the proof is to establish the correct dependence on
the parameter $\gamma$ since it is not directly related to the coin estimation problem.
We elaborate on it in \Cref{rem:coin range of parameters}.

\begin{remark}
    Our construction, 
    combined with the non-replicable version of the coin estimation problem,
    can be used to simplify the construction of the non-replicable $Q$-estimation lower bound from \citet{gheshlaghi2013minimax}.
\end{remark}

\subsubsection{The Replicable Coin Estimation Problem}\label{sec:coin problem}
Formally, the estimation problem, without the replicability requirement,
is defined as follows.
\begin{prob}[Multiple Coin Problem]\label{prob:multi coin}
  Fix $q, \varepsilon, \delta\in (0, 1)^3$ such that $q-\varepsilon \in (\nicefrac12, 1)$.
  Given sample access to $N$ independent coins each with a bias of either $q$ or $q-\varepsilon$,
  determine the bias of every coin
  with confidence at least $1-\delta$.
\end{prob}

We now informally state our main result for the multiple coin estimation problem,
which could be useful in deriving replicability
lower bounds beyond the scope of our work.
See \Cref{thm:replicable multi bias lower bound} for the formal statement.
Intuitively,
this result generalizes \Cref{thm:SQ-lowerbound} to multiple instances.
\begin{restatable}[Informal]{thm}{rMultiCoinLowerBoundInformal}\label{thm:replicable multi bias lower bound informal}
    Suppose $\mscr A$
    is a locally random $\rho$-replicable algorithm
    for the multiple coin problem
    with a constant probability of success.
    Then, the sample complexity of $\mscr A$ is at least
    \[
      \Omega\left( \frac{N^3 q(1-q)}{\varepsilon^2 \rho^2} \right).
    \]
\end{restatable}

Recall Yao's min-max principle \citep{yao1977probabilistic},
which roughly states that the expected cost of a randomized algorithm on its worst-case input
is at least as expensive as the expected cost of any deterministic algorithm on random inputs chosen from some distribution.
It is not clear how to apply Yao's principle directly, but we take inspiration from its essence
and reduce the task of reasoning about a randomized algorithm with shared internal randomness
to reasoning about a deterministic one with an additional
 layer of external randomness on top of the 
random flips of the coins.  

\iffalse
On a high level,
we can think of a $\rho$-replicable algorithm $\mscr A(\bar x; \bar r)$ for the multiple coin estimation problem
as a distribution over deterministic algorithms $g_{\bar r}(\bar x)$ with no internal randomness.
We argue that at least a constant fraction of $g_{\bar r}$'s will satisfy the same guarantees as $\mscr A$
and so it suffices to reason about these ``nice'' $g_{\bar r}$'s.
For each fixed $\bar r$,
$g_{\bar r}$ is independent across different coins and different executions,
hence we can apply a union bound argument.
Thus instead of reasoning about an algorithm with shared internal randomness across multiple coins,
we can reason about deterministic algorithms across a single coin.
\fi

Consider now a deterministic algorithm $g$ for distinguishing the bias of a single coin
where the input bias is chosen uniformly in $[q-\varepsilon, q]$.
That is,
we first choose $p\sim U[q-\varepsilon, q]$,
then provide i.i.d. samples from $\Be(p)$ to $g$.
We impose some boundary conditions: if $p=q-\varepsilon$,
$g$ should output ``-'' with high probability
and if $p=q$,
the algorithm should output ``+'' with high probability.
We show that the probability of $g$ outputting ``+'' varies smoothly with respect to the bias of the input coin.
Thus, there is an interval $I\sset (q-\varepsilon, q)$
such that $g$ outputs ``-'' or ``+'' with almost equal probability
and so the output of $g$ is inconsistent across two executions with constant probability when $p$ lands in this interval.
By the choice of $p\sim U[q-\varepsilon, q]$,
if $\ell(I)$ denotes the length of $I$,
then the output of $g$ is inconsistent across two executions
with probability at least $\Omega(\nicefrac{\ell(I)}\varepsilon)$.
Quantifying $\ell(I)$ and rearranging yields the lower bound for a single coin.

For the case of $N$ independent coins,
we use the pigeonhole principle to reduce the argument to the case of a single coin.
The formal statement and proof of \Cref{thm:replicable multi bias lower bound informal}
is deferred to \Cref{apx:coin problem}.

\iffalse
\begin{remark}
    \Cref{thm:replicable multi bias lower bound informal} does not insist that the algorithm draws the same number of samples for every coin.
\end{remark}
\fi

\begin{remark}\label{rem:coin range of parameters}
    The lower bound from \citet{impagliazzo2022reproducibility} for the single-coin estimation problem holds for the regime $q, q-\varepsilon\in (\nicefrac14, \nicefrac34)$.
    We remove this constraint by analyzing the dependence of the lower bound on $q$.
    When reducing $Q$-function estimation to the multiple coin problem,
    the restricted regime yields a lower bound proportional to $(1-\gamma)^{-2}$.
    In order to derive the stronger lower bound of $(1-\gamma)^{-3}$,
    we must be able to choose $q\approx \gamma$ which can be arbitrarily close to 1.
\end{remark}

In \Cref{sec:tv ind section},
we show that allowing for non-locally random algorithms
enables us to shave off a factor of $N$ in the sample complexity.
We also conjecture that this upper bound is tight.

\begin{conj}\label{conj:replicable randomized coin lower bound}
    Suppose $\mscr A(\bar c^{(1)}, \dots, \bar c^{(N)}; \bar r)$
    is a randomized $\rho$-replicable algorithm 
    for the multiple coin problem
    and has a constant probability of success.
    Then, the sample complexity of $\mscr A$ is at least
    \[
      \Omega\left( \frac{N^2 q(1-q)}{\varepsilon^2 \rho^2} \right).
    \]
\end{conj}

\subsubsection{A Lower Bound for Replicable \texorpdfstring{$Q$}{Q}-Function Estimation}\label{sec:q lower bound}
\begin{figure}[h]
  \centering
  \includegraphics[width=0.85\textwidth]{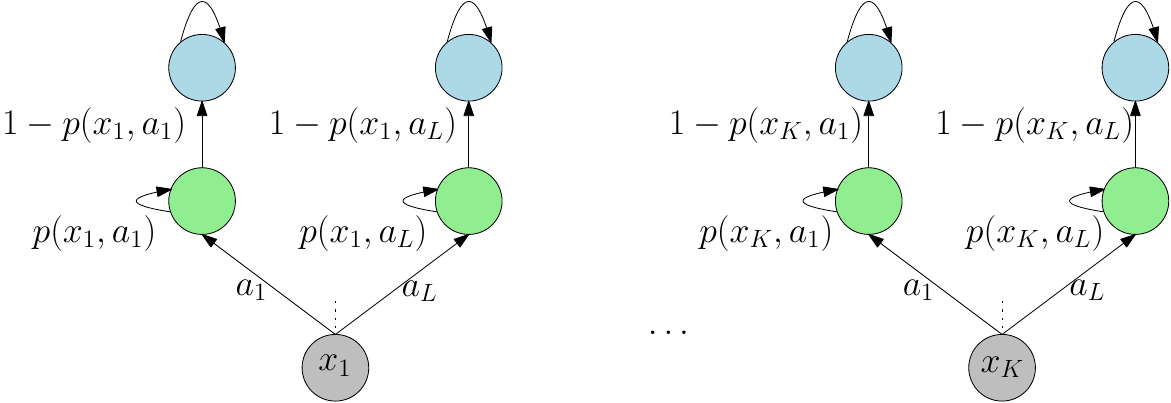}
  \caption{The class of MDPs considered to prove the lower bound in \Cref{thm:replicable Q estimation lower bound}.}
  \label{fig:replicable Q estimation lower bound}
\end{figure}

We now present the MDP construction that achieves the desired sample complexity lower
bound.
We define a family of MDPs $\mbb M$ as depicted in \Cref{fig:replicable Q estimation lower bound}.
This particular construction was first presented by \citet{mannor2004sample}
and generalized by \citet{gheshlaghi2013minimax,feng2019does}.
%\citet{feng2019does}.

Any MDP $M\in \mbb M$ is parameterized by positive integers $K_M, L_M$,
and some $p_M^{(k, \ell)}\in [0, 1]$ for $k\in [K_M], \ell\in [L_M]$.
The state space of $M$ is the disjoint union\footnote{Denoted by $\sqcup$.} of three sets
$\mcal S = \mcal X \sqcup \mcal Y\sqcup \mcal Z$,
where $\mcal X$ consists of $K$ states $\set{x_1, \dots, x_K}$
and each of them has $L$ available actions $\set{a_1, \dots, a_L} =: \mcal A$.
All states in $\mcal Y, \mcal Z$ have a single action that the agent can take.
Remark that each $M\in \mbb M$ has
$N = \sum_{s\in S} \card{\mcal A^s} = 4K_M L_M$.

For $x\in \mcal X$,
by taking action $a\in \mcal A$,
the agent transitions to a state $y(x, a)\in \mcal Y$ with probability 1.
Let $p_M(x_k, a_\ell) := p_M^{(k, \ell)}$.
For state $y(x, a)\in \mcal Y$,
we transition back to $y(x, a)$ with probability $p_M(x, a)$
and to $z(x, a)\in \mcal Z$ with probability $1-p_M(x, a)$.
Finally,
the agent always returns to $z(x, a)$ for all $z(x, a)\in \mcal Z$.
The reward function $r_M(s, a) = 1$ if $s \in \mcal X\cup \mcal Y$ and is 0 otherwise.
We remark that for every $x\in \mcal X, a\in \mcal A$,
its $Q^\star$ function can be computed in closed form
by solving the Bellman optimality equation
\begin{align*}
  Q_M^\star(x, a)
  &= 1 + \gamma \left[ p_M(x, a)\cdot Q_M^\star(x, a) + (1-p_M(x, a))\cdot 0 \right]
  = \frac1{1-\gamma p_M(x, a)}.
\end{align*}

Recall we write $N := \sum_{s\in \mcal S} \card{\mcal A^s}$.
to denote the total number of state-action pairs.
Our main result
in this section is the following.
\begin{restatable}{thm}{rQLowerBound}\label{thm:replicable Q estimation lower bound}
    Let $\rho, \varepsilon\in (0, 1)^2$,
    $\gamma\in (\nicefrac12, 1)$,
    and $\delta = \nicefrac14$.
    Suppose $\mscr A$ is a locally random $\rho$-replicable algorithm
    that returns an estimate $\wh Q$ for any MDP
    with discount factor $\gamma$
    such that
    \mbox{$\abs{\wh Q(s, a) - Q^\star(s, a)} \leq \varepsilon$}
    with probability at least $1-\nicefrac\delta6$
    for each $s\in \mcal S, a\in \mcal A^s$.
    Then $\mscr A$ has a sample complexity of at least
    \[
        \Omega\left( \frac{N^3}{(1-\gamma)^3 \varepsilon^2 \rho^2} \right).
    \]
\end{restatable}

\begin{remark}
    If \Cref{conj:replicable randomized coin lower bound} holds,
    we obtain a sample complexity lower bound of
    \[
        \Omega\left( \frac{N^2}{(1-\gamma)^3 \varepsilon^2 \rho^2} \right)
    \]
    for general randomized $\rho$-replicable algorithms for $Q$ estimation.
\end{remark}

On a high level,
we argue that a locally random $\rho$-replicable algorithm $\mscr A$ for estimating the $Q$ function of arbitrary MDPs up to accuracy $\varepsilon \approx \nicefrac{\varepsilon_0}{(1-\gamma)^2}$
yields a locally random $\rho$-replicable algorithm for the multiple coin problem (cf. \Cref{prob:multi coin})
with tolerance approximately $\varepsilon_0 \approx (1-\gamma)^2\varepsilon$
when we choose $q\approx \gamma$ in \Cref{thm:replicable multi bias lower bound informal}.
We can then directly apply \Cref{thm:replicable multi bias lower bound informal} to conclude the proof.
See \Cref{apx:replicable Q lower bound} for details.

\subsubsection{A Lower Bound for Replicable Policy Estimation}
Having established the lower bound for locally random replicable $Q$-function
estimation, we now present our lower bound for deterministic replicable policy 
estimation.  
We argue that a deterministic $\rho$-replicable algorithm for optimal policy estimation
yields a locally random $\rho$-replicable algorithm for optimal $Q$-function estimation
after some post-processing that has sample complexity $\tilde o\left(N^3/\varepsilon^2\rho^2(1-\gamma)^3 \right)$.
It follows that the sample complexity lower bound we derived for $Q$-function estimation holds for policy estimation as well.

In order to describe the post-processing step,
we employ a locally random replicable rounding algorithm (cf. \Cref{thm:replicable rounding}) that is
provided in \citet{esfandiari2023replicable}.
\iffalse
and has sample
complexity $O(\log(1/\delta)/(\varepsilon^2\rho^2))$ (cf. \Cref{thm:replicable mean estimation}), where $\delta$ is the confidence of the estimation, $\varepsilon$ is
its accuracy, and $\rho$ is the replicability parameter.
% the replicable SQ oracle from \citet{impagliazzo2022reproducibility}
% which we state below for convenience.
\fi
Intuitively,
we show that estimating the value function $V^\pi$ of $\pi$
reduces to estimating the optimal $Q$-function of some single-action MDP.
Given such an estimate $\hat V^\pi$,
we can then estimate $Q^\pi$ using the simple sample mean query 
given sufficient samples from the generative model. 
Lastly,
the locally random replicable rounding subroutine from \Cref{thm:replicable mean estimation}
is used as a post-processing step.

We now state the formal lower bound regarding the sample complexity of deterministic replicable policy estimation. 
Its proof follows by combining the $Q$-function estimation lower bound
and the reduction we described above.
For the full proof,
see \Cref{apx:replicable policy lower bound}.

\begin{restatable}{thm}{rPolicyLowerBound}\label{thm:replicable policy lower bound}
  Let $\varepsilon, \rho\in (0, 1)^2$ and $\delta = \nicefrac14$.
  Suppose $\mscr A$ is a deterministic $\rho$-replicable algorithm that outputs a randomized policy $\pi$
  such that
  $\abs{V^\pi(s) - V^\star(s)} \leq \varepsilon$
  with probability at least $1-\nicefrac\delta{12}$
  for each $s\in \mcal S$.
  Then $\mscr A$ has a sample complexity of at least
  \[
    \Omega\left( \frac{N^3}{(1-\gamma)^3 \varepsilon^2 \rho^2} \right).
  \]
\end{restatable}

\begin{remark}
    If \Cref{conj:replicable randomized coin lower bound} holds,
    we obtain a sample complexity lower bound of
    \[
        \Omega\left( \frac{N^2}{(1-\gamma)^3 \varepsilon^2 \rho^2} \right)
    \]
    for general randomized $\rho$-replicable algorithms for policy estimation.
\end{remark}

\subsection{The Coin Problem}\label{apx:coin problem}
As mentioned before,
estimating the bias of a coin is an elementary statistical problem.
In order to establish our lower bounds,
we first aim to understand the sample complexity of locally random algorithms that replicably estimate the bias of multiple coins simultaneously.

\citet{impagliazzo2022reproducibility} explored the single coin version of the problem
and established a version of the following result when the biases $q, q-\varepsilon$ of the coins in question lie in the interval $(\nicefrac14, \nicefrac34)$.
We state and prove the more general result when we allow $q$ to be arbitrarily close to $1$.
\begin{thm}[SQ-Replicability Lower Bound; \citep{impagliazzo2022reproducibility}]\label{thm:SQ-lowerbound}
    Fix $q, \varepsilon, \rho\in (0, 1)^3$ such that $q-\varepsilon \in (\nicefrac12, 1)$
    and $\delta = \nicefrac14$.
    Let $\mscr A(\bar c; \bar r)$
    be an algorithm for the (single) coin problem.
    
    Suppose $\mscr A$ satisfies the following:
    \begin{enumerate}[(i)]
      \item $\mscr A$ outputs $\set{0, 1}$ 
        where 1 indicates its guess
        that the bias of the coin which generated $\bar c$ is $q$.
      \item \label{item:one-coin distributional replicability} $\mscr A$ is $\rho$-replicable even when its samples are drawn from coins
        with bias $p\in [q-\varepsilon, q]$
        for all $i\in [N]$.
      \item \label{item:one-coin boundary conditions} If $p\in \set{q-\varepsilon, q}$,
        then $\mscr A$ correctly guesses the bias of the $i$-th coin with probability at least $1-\nicefrac{\delta}{6}$.
    \end{enumerate}

    Then the sample complexity of $\mscr A$ is at least
    \[
      \Omega\left( \frac{q(1-q)}{\varepsilon^2 \rho^2} \right).
    \]
\end{thm}

We follow a similar proof process as \citet{impagliazzo2022reproducibility}
towards a lower bound for multiple coins.
In particular,
we begin with the two following lemmas adapted from \citet{impagliazzo2022reproducibility}.

\begin{lem}[\citep{impagliazzo2022reproducibility}]\label{lem:acc derivative}
  Let $g: \set{0, 1}^m \to \set{0, 1}$ be a boolean function.
  Suppose the input bits are independently sampled from $\Be(p)$ for some parameter $p\in[0, 1]$
  and let \mbox{$\Acc: [0, 1]\to [0, 1]$} be given by
  \[
    \Acc(p) := \P_{\bar x\sim_{i.i.d.} \Be(p)}\set{g(\bar x) = 1}.
  \]
  Then $\Acc$ is differentiable on $(0, 1)$
  and for all $p\in (0, 1)$,
  \[
    \Acc'(p) \leq \sqrt{\frac{m}{p(1-p)}}.
  \]
\end{lem}

\begin{pf}[\Cref{lem:acc derivative}]
  Fix $p\in (0, 1)$ and suppose $\bar x\sim_{i.i.d.} \Be(p)$.
  Define
  \[
    a_k := \P\set*{g(\bar x) = 1 \,\middle|\, \sum_{i=1}^m x_i = k}.
  \]
  Then
  \[
    \Acc(p)
    = \sum_{k=1}^m a_k \binom{m}{k} p^k (1-p)^{m-k}.
  \]
  In particular,
  $\Acc$ is differentiable.

  By computation,
  \begin{align*}
    \Acc'(p)
    &= \sum_{k=1}^m a_k \binom{m}{k} \left[ kp^{k-1} (1-p)^{m-k} - (m-k)p^k(1-p)^{m-k-1} \right] \\
    &= \sum_k a_k \binom{m}{k} p^k(1-p)^{m-k} \left[ \frac{k}p - \frac{m-k}{1-p} \right] \\
    &= \sum_k a_k \binom{m}{k} p^k(1-p)^{m-k}\cdot \frac{k(1-p) - (m-k)p}{p(1-p)} \\
    &= \sum_k a_k \binom{m}{k} p^k(1-p)^{m-k}\cdot \frac{k - mp}{p(1-p)} \\
    &\leq \sum_k \binom{m}{k} p^k(1-p)^{m-k}\cdot \frac{\abs{k - mp}}{p(1-p)} &&a_k\in [0, 1] \\
    &= \frac1{p(1-p)} \E\left[\, \abs{X - \E[X]} \,\right] &&X\sim \Bin(m, p) \\
    &\leq \frac1{p(1-p)} \sqrt{\Var[X]} \\
    &= \frac1{p(1-p)} \sqrt{mp(1-p)} \\
    &= \sqrt{\frac{m}{p(1-p)}}. \qedhere
  \end{align*}
\end{pf}

In the following arguments,
we need to reason about probabilistic events with multiple sources of randomness.
For the sake of clarity,
we use the formal definitions of a probability space and random variable.
Specifically,
let $(W, \mcal F, \P)$ be an underlying probability space
where $W$ is some sample space,
$\mcal F\sset 2^W$ is a $\sigma$-algebra,
and $\P$ is a probability measure.
A random variable is simply a real-valued function $W\to \R$ from the sample space.

Define $C_p := \Be(p)$.
Moreover,
define $C_{-} := \Be(q-\varepsilon)$
and $C_{+} :=\Be(q)$
to be the distributions of the possible biased coins. 
\begin{lem}[\citep{impagliazzo2022reproducibility}]\label{lem:replicable coin lowerbound}
  Fix $q, \varepsilon\in (0, 1)$ such that $q-\varepsilon \in (\nicefrac12, 1)$
  and $\delta = \nicefrac14$.
  Suppose $g: \set{0, 1}^m\to \set{0, 1}$ is a boolean function satisfying the following:
  \begin{enumerate}[(i)]
    \item For $\bar x\sim_{i.i.d.} C_{-}$,
        $\P\set{w: g(\bar x(w)) = 0} \geq 1-\delta$.
    \item For $\bar x\sim_{i.i.d.} C_{+}$,
        $\P\set{w: g(\bar x(w)) = 1} \geq 1-\delta$.
  \end{enumerate}
  Then for $p\sim U[q-\varepsilon, q]$
  and $\bar x^{(1)}, \bar x^{(2)} \sim_{i.i.d.} C_p$,
  \[
    \P\set{w: g(\bar x^{(1)}(w)) \neq g(\bar x^{(2)}(w))}
    \geq \Omega\left( \frac{\sqrt{q(1-q)}}{\varepsilon \sqrt{m}} \right).
  \]
  In other words,
  we require
  \[
    m \geq \Omega\left( \frac{q(1-q)}{\varepsilon^2 \rho^2} \right)
  \]
  if we wish to reduce the probability above to at most $\rho$.
\end{lem}

We should interpret the input of the function $g$ from \Cref{lem:replicable coin lowerbound}
as $m$ realizations of coin flips from the same coin
and the output of $g$ as its guess
whether the bias of the coin that generated the realizations
is $q$.
\Cref{lem:replicable coin lowerbound} states that if the function $g$ is ``nice'',
i.e. is able to distinguish the two coins $C_{-}, C_{+}$
with some fixed confidence $1-\delta$,
then the same function is not replicable with probability at least $\Omega(\nicefrac{\sqrt{q(1-q)}}{\varepsilon \sqrt m})$
when the bias is chosen uniformly randomly in the interval $[q-\varepsilon, q]$.
Let us view an arbitrary $\rho$-replicable algorithm $\mscr A(\bar c; \bar r)$
as a distribution over functions $g_{\bar r}(\bar c)$.
\citet{impagliazzo2022reproducibility} argued that at least a constant fraction of $g_{\bar r}$ is nice,
leading to \Cref{thm:SQ-lowerbound}.
Unfortunately,
this argument does not extend trivially to the case of multiple coins.
However,
we present an extension for the special case when $\mscr A$ is locally random.

\begin{pf}[\Cref{lem:replicable coin lowerbound}]
  Let $g: \set{0, 1}^m\to \set{0, 1}$ be a boolean function
  that satisfies the condition of \Cref{lem:replicable coin lowerbound}.
  By \Cref{lem:acc derivative},
  \[
    \Acc(p) := \P_{\bar c\sim_{i.i.d.} C_p} \set{w: g(\bar c(w)) = 1}
  \]
  is differentiable (continuous)
  and for every $p\in [q-\varepsilon, q]$,
  \[
    \Acc'(p) \leq \sqrt{\frac{m}{q(1-q)}}.
  \]
  This is because $\frac1{p(1-p)}$ is non-increasing on $(\nicefrac12, 1)$.
  In particular,
  $\Acc(p)$ is $\sqrt{\frac{m}{q(1-q)}}$-Lipschitz over the interval $[q-\varepsilon, q]$.

  Now,
  $\Acc(q-\varepsilon) \leq \delta < \nicefrac14$
  and $\Acc(q) \geq 1-\delta > \nicefrac34$,
  thus by the intermediate value theorem from elementary calculus,
  there is some $q_0\in (q-\varepsilon, q)$ such that $\Acc(q_0) = \nicefrac12$.
  It follows by the Lipschitz condition
  and the mean value theorem
  that there is an interval $I$ of length $\Omega(\sqrt{\nicefrac{q(1-q)}{m}})$ around $q_0$
  so that for all $p\in I$,
  \[
    \Acc(p)\in \left( \frac13, \frac23 \right).
  \]

  For two independent sequences of i.i.d. samples,
  say $\bar x^{(1)}, \bar x^{(2)}$,
  we know that $g(\bar x^{(1)})$ and $g(\bar x^{(2)})$ are independent.
  Thus for $p\in I$,
  there is a $2\Acc(p)(1-\Acc(p)) > \nicefrac49$
  probability that \mbox{$g(\bar x^{(1)}) \neq g(\bar x^{(2)})$}.

  Then for $p\sim U[q-\varepsilon, q]$
  and $\bar x^{(1)}, \bar x^{(2)} \sim_{i.i.d.} C_p$,
  \begin{align*}
    &\P\set*{w: g(\bar x^{(1)}(w)) \neq g(\bar x^{(2)}(w))} \\
    &\geq \P\set*{w: g(\bar x^{(1)}(w)) \neq g(\bar x^{(2)}(w)) \,\middle|\, p(w)\in I}\cdot \P\set*{w: p(w)\in I} \\
    &\geq \frac49\cdot \Omega\left( \frac{\sqrt{\nicefrac{q(1-q)}{m}}}{2\varepsilon} \right) &&p\sim U\left[ q-\varepsilon, q \right] \\
    &= \Omega\left( \frac{\sqrt{q(1-q)}}{\varepsilon \sqrt{m}} \right). \qedhere
  \end{align*}
\end{pf}

\Cref{lem:replicable coin lowerbound} enables us to prove \Cref{thm:SQ-lowerbound}.

\begin{pf}[\Cref{thm:SQ-lowerbound}]
    Let $G_{-}$ be the event that $\mscr A$ correctly guess the bias of the coin given $p = q-\varepsilon$
    and similarly for $G_{+}$.
    Thus
    \begin{align*}
        G_{-}
        & := \set*{w: \mscr A(\bar c(w); \bar r(w)) = 0} 
        && p = q-\varepsilon \\
        G_{+}
        & := \set*{w: \mscr A(\bar c(w); \bar r(w)) = 1}
        && p = q.
    \end{align*}
    Here the randomness lies only in the samples $\bar c\sim_{i.i.d.} C_{p}$
    and the uniformly random $\bar r$.
    By \ref{item:one-coin boundary conditions},
    \begin{align*}
        \frac\delta{3\cdot 2}
        &\geq \sum_{\bar r} \P(G_{-}^c\mid \bar r)\cdot \P(\bar r) \\
        &\geq \E_{\bar r} [\P(G_{-}^c\mid \bar r)].
    \end{align*}
    Thus Markov's inequality tells us that choosing $\bar r$ uniformly at random satisfies
    \[
      \P(G_{-}^c\mid \bar r) \leq \delta
    \]
    with probability at least $1-\nicefrac1{(3\cdot 2)}$
    and similarly for $G_+$.

    By a union bound,
    choosing $\bar r$ uniformly at random means $G_{\pm}$ both occur
    with probability at least $\nicefrac23$.
    Let us write $\bar r\in G := G_+\cap G_-$ as a shorthand for indicating the random draw $\bar r$ satisfies both $G_\pm$.
    
    Fix $\bar r^\star\in G$ to be any particular realization
    and observe that $g := \mscr A(\, \cdot\, ; \bar r^\star)$
    satisfies the condition of \Cref{lem:replicable coin lowerbound}.
    By \Cref{lem:replicable coin lowerbound},
    if $p\sim U[q-\varepsilon, q]$,
    and $\bar c^{(1)}, \bar c^{(2)} \sim_{i.i.d.} C_p$,
    \begin{align*}
        &\P\set*{w: \mscr A(\bar c^{(1)}(w); r(w)) \neq \mscr A(\bar c^{(2)}(w); r(w))} \\
        &= \sum_{\bar r} \P\set*{w: \mscr A(\bar c^{(1)}(w); r(w)) \neq \mscr A(\bar c^{(2)}(w); r(w)) \mid \bar r}\cdot \P(\bar r) \\
        &\geq \sum_{\bar r\in G} \P\set*{w: \mscr A(\bar c^{(1)}(w); r(w)) \neq \mscr A(\bar c^{(2)}(w); r(w)) \mid \bar r}\cdot \P(\bar r) \\
        &\geq \frac23\cdot \Omega\left( \frac{\sqrt{q(1-q)}}{\varepsilon \sqrt{m}} \right) \\
        &= \Omega\left( \frac{\sqrt{q(1-q)}}{\varepsilon \sqrt{m}} \right).
    \end{align*}
    We remark that in the derivation above,
    a crucial component is the fact that outputs of $\mscr A$ across two runs are conditionally independent given the internal randomness.

    Since \ref{item:one-coin distributional replicability} also holds when $p$ is chosen uniformly at random
    in the interval $[q-\varepsilon, q]$,
    it follows that $\Omega( \nicefrac{\sqrt{q(1-q)}}{\varepsilon \sqrt{m}} )\leq \rho$ is a lower bound for the replicability parameter.
    The total sample complexity is thus at least
    \[
      \Omega\left( \frac{q(1-q)}{\varepsilon^2 \rho^2} \right). \qedhere
    \]
\end{pf}

We now use \Cref{thm:SQ-lowerbound} to prove the formal statement of \Cref{thm:replicable multi bias lower bound informal},
a samples complexity lower bound
for locally random algorithms
that replicably distinguish the biases of $N$ independent coins
$C^{(1)}, \dots, C^{(N)}$,
assuming each of which is either $C_{-}$ or $C_{+}$.
The lower bound for locally random replicable $Q$-estimation follows.
We formally state \Cref{thm:replicable multi bias lower bound informal} below.
\begin{restatable}[\Cref{thm:replicable multi bias lower bound informal}; Formal]{thm}{rMultiCoinLowerBound}\label{thm:replicable multi bias lower bound}
    Fix $q, \varepsilon, \rho\in (0, 1)^3$ such that $q-\varepsilon \in (\nicefrac12, 1)$
    and $\delta = \nicefrac14$.
    Let $\mscr A = (\mscr A^{(1)}, \dots, \mscr A^{(N)})$
    be a locally random algorithm for the multiple coin problem
    where $\mscr A^{(i)}(\bar c^{(1)}, \dots, \bar c^{(N)}; \bar r^{(i)})$ 
    is the output for the $i$-th coin
    that draws internal randomness independently from other coins.
    
    Suppose $\mscr A$ satisfies the following:
    \begin{enumerate}[(i)]
      \item $\mscr A^{(i)}$ outputs $\set{0, 1}$ 
        where 1 indicates its guess
        that the bias of the coin which generated $\bar c^{(i)}$ is $q$.
      \item \label{item:distributional replicability} $\mscr A$ is $\rho$-replicable even when its samples are drawn from coins
        where $p^{(i)}\in [q-\varepsilon, q]$
        for all $i\in [N]$.
      \item \label{item:boundary conditions} If $p^{(i)}\in \set{q-\varepsilon, q}$,
        then $\mscr A$ correctly guesses the bias of the $i$-th coin with probability at least $1-\nicefrac{\delta}{6}$.
    \end{enumerate}

    Then the sample complexity of $\mscr A$ is at least
    \[
      \Omega\left( \frac{N^3 q(1-q)}{\varepsilon^2 \rho^2} \right).
    \]
\end{restatable}
We remark that the scaling with respect to $q$ is important in order to establish a strong lower bound with respect to the parameter $(1-\gamma)^{-1}$.

\begin{pf}[\Cref{thm:replicable multi bias lower bound}]
    First,
    we remark that the $i$-th output of $\mscr A$ across two runs
    is independent of other outputs
    since both the coin flips
    and internal randomness are drawn independently per coin.
    Thus we may as well assume that $\mscr A^{(i)}$ only reads the coin flips from the $i$-th coin.

    Let $\rho_i, i\in [N]$ be the probability that the output of $\mscr A^{(i)}$
    is inconsistent across two executions
    when the bias $p^{(i)}$ is chosen uniformly in $[q-\varepsilon, q]$.
    We claim that there are at least $\nicefrac{N}2$ indices $i\in [N]$
    such that $\rho_i \leq \nicefrac{\rho}{N}$.
    Indeed,
    by the independence due to the local randomness property,
    we have
    \begin{align*}
        \rho
        &\geq 1 - \prod_{i\in [N]} \left( 1-\rho_i \right) \\
        &\geq 1 - \exp\left( -\sum_i \rho_i \right) &&1+x\leq e^x \\
        &\geq \sum_i \frac{\rho_i}2. &&e^{-x} \leq 1-\frac{x}2, x\in [0, 1]
    \end{align*}
    Suppose towards a contradiction that more than $\nicefrac{N}2$ indices $i\in [N]$
    satisfy $\rho_i > \nicefrac\rho{N}$.
    But then
    \[
        \sum_i \frac{\rho_i}2
        > \frac{N}2\cdot 2 \frac\rho{N}
        = \rho,
    \]
    which is a contradiction.
    
    Let $I\sset [N]$ be the indices of the coins
    such that $\rho_i \leq \nicefrac\rho{N}$.
    For each $i\in I$,
    $\mscr A^{(i)}$ satisfies the conditions for \Cref{thm:SQ-lowerbound}.
    Thus $\mscr A^{(i)}$ has sample complexity at least
    \[
        \Omega\left( \frac{q(1-q)}{\varepsilon^2 (\rho/N)^2} \right).
    \]

    It follows that the total sample complexity is at least
    \[
      \Omega\left( \frac{N}2\cdot \frac{q(1-q)}{\varepsilon^2 (\rho/N)^2} \right)
      = \Omega\left( \frac{N^3 q(1-q)}{\varepsilon^2 \rho^2} \right). \qedhere
    \]
\end{pf}

\subsection{Replicable \texorpdfstring{$Q$}{Q}-Function Estimation Lower Bound}\label{apx:replicable Q lower bound}
In this section,
we restate and prove \Cref{thm:replicable Q estimation lower bound},
a lower bound on locally random replicable $Q$-function estimation.
\rQLowerBound*

\begin{pf}[\Cref{thm:replicable Q estimation lower bound}]
    Choose $q := \nicefrac{4\gamma - 1}{3\gamma}$
    and $\varepsilon_0\in (0, \nicefrac{(1-\gamma)}{\gamma})$
    such that $q-\varepsilon_0\in (\nicefrac12, 1)$.
    Let $C^{(i)}\sim \Be(p^{(i)})$ be Bernoulli random variables (biased coins) for $i\in [N]$
    where $p^{(i)}\in [q-\varepsilon_0, q]$.
    
    To see the reduction,
    first choose any $K, L\in \Z_+$ such that $KL = N$
    and let $M\in \mbb M$ be such that the state action space has cardinality $4N$
    as in the construction in \Cref{fig:replicable Q estimation lower bound}.
    We identity each $i\in [N]$ with a pair $(x, a)\in \mcal X\times \mcal A$
    and write $i_{x, a}$ to be the index corresponding to the pair $(x, a)$,
    i.e. $p(x, a) = p^{(i_{x, a})}$.
    For each state $y(x, a)\in \mcal Y$,
    we flip the coin $C^{(i_{x, a})}\sim \Be(p(x, a)))$
    and transition back to $y(x, a)$ if the coin lands on 1
    and to $z(x, a)\in \mcal Z$ if it lands on 0.
    By construction,
    \[
        Q_M^\star(x, a)
        = \frac1{1-\gamma p_M(x, a)}.
    \]

    Let us compare the absolute difference of $Q^\star(x, a)$ when $p(x, a) = q$ versus when $p(x, a) = q-\varepsilon_0$.
    By computation,
    \begin{align*}
        \frac1{1-\gamma q} - \frac1{1-\gamma (q-\varepsilon_0)}
        &=  \frac{\gamma\varepsilon_0}{[1-\gamma(q-\varepsilon_0)][1-\gamma q]}.
    \end{align*}
    Then the following inequalities hold.
    \begin{align*}
        q
        &\geq \frac23 \\
        1-q
        &= \frac{1 - \gamma}{3\gamma} \\
        &\geq \frac23 (1-\gamma) &&\gamma > \frac12 \\
        1 - \gamma q
        &= \frac43 (1 - \gamma) \\
        1 - \gamma (q-\varepsilon_0)
        &= \frac43(1-\gamma) + \gamma \varepsilon_0 \\
        &\leq \frac43(1-\gamma) + (1-\gamma) &&\varepsilon_0 < \frac{1-\gamma}\gamma \\
        &= \frac73 (1-\gamma).
    \end{align*}
    It follows that
    \begin{align*}
        \abs*{\frac1{1-\gamma q} - \frac1{1-\gamma (q-\varepsilon_0)}}
        &\geq \frac{9 \gamma \varepsilon_0}{28 (1-\gamma)^2} \\
        &\geq \frac{3 \varepsilon_0}{14 (1-\gamma)^2} &&\gamma\geq \frac23 \\
        &=: \varepsilon.
    \end{align*}

    Suppose we run a locally random algorithm that replicably estimates $Q_M^\star(x, a)$ to an accuracy of $\nicefrac{\varepsilon}{3}$.
    Then we are able to distinguish whether the biases of the coins $C^{(i_{x, a})}$ is either $q-\varepsilon_0$ or $q$.
    In addition,
    the internal randomness consumed in the estimation of $C^{(i_{x, a})}$
    comes only from the internal randomness used to estimate $Q^\star(x, a)$.
    Hence the scheme we described is a locally random algorithm for the multiple coin problem.
    Finally,
    the scheme is replicable even when the biases are chosen in the interval $[q-\varepsilon_0, q]$.
    
    By \Cref{thm:replicable multi bias lower bound},
    the following sample complexity lower bound holds for $\mscr A$:
    \begin{align*}
        \Omega\left( \frac{N^3 q(1-q)}{\varepsilon_0^2 \rho^2} \right)
        &= \Omega\left( \frac{N^3 (1-\gamma)}{[(1-\gamma)^2\varepsilon]^2 \rho^2} \right)
        = \Omega\left( \frac{N^3}{(1-\gamma)^3\varepsilon^2 \rho^2} \right).  \qedhere
    \end{align*}
\end{pf}

\subsection{Replicable Policy Estimation Lower Bound}\label{apx:replicable policy lower bound}
In order to prove a lower bound on locally random replicable policy estimation,
we first describe a locally random replicable algorithm 
that estimates the state-action function $Q^\pi$ for a given (possibly randomized) policy $\pi$.
Recall $N := \sum_{s\in \mcal S} \card{\mcal A^s}$ denotes the number of state-action pairs.
\begin{lem}\label{lem:replicable policy to Q}
  Let $\varepsilon, \rho\in (0, 1)$ and $\delta\in (0, \nicefrac\rho3)$.
  Suppose $\pi$ is an explicitly given randomized policy.
  There is a locally random $\rho$-replicable algorithm
  that outputs an estimate $\wh Q^\pi$ of $Q^\pi$
  such that with probability at least $1-\delta$,
  $\norm*{\wh Q^\pi - Q^\pi}_\infty \leq \varepsilon$.
  Moreover,
  the algorithm has sample complexity
  \[
    \tilde O\left( \left( \frac{\card{\mcal S} N^2}{(1-\gamma)^3 \varepsilon^2 \rho^2} + \frac{N^3}{(1-\gamma)^2 \varepsilon^2 \rho^2} \right)\log\frac1\delta \right)
    = \tilde o\left( \frac{N^3}{(1-\gamma)^3 \varepsilon^2 \rho^2} \log\frac1\delta \right).
  \]
\end{lem}

\begin{pf}[\Cref{lem:replicable policy to Q}]
  Let $\pi: \mcal S\to \Delta(\mcal A)$ be explicitly given.
  First,
  we construct a single-action MDP $M' = (\mcal S, s_0, \set{0}, P', r', \gamma)$ whose optimal (and only) state-action function $Q'$ is precisely the value function $V^\pi$ of $\pi$.

  Let the state space $\mcal S$ and initial state $s_0$ remain the same
  but replace the action space with a singleton ``0''.
  We identify each new state-action pair with the state since only one action exists.
  Next,
  define the transition probability
  \[
    P'(s_1\mid s_0)
    := \sum_{a\in \mcal A^{s_0}} \pi(s_0, a)\cdot P(s_1\mid s_0, a).
  \]
  We can simulate a sample from $P'(\cdot\mid s_0)$ by sampling (for free) from the policy,
  say $a\sim \pi(s_0)$,
  and then sampling from the generative model $s_1\sim P(\cdot\mid s_0, a)$.
  Note this costs a single sample from the generative model.
  In addition,
  define the reward function as the expected reward at state $s_0$
  \[
    r'(s_0)
    := \sum_{a\in \mcal A^{s_0}} \pi(s_0, a)\cdot r(s_0, a).
  \]
  This value can be directly computed given $\pi, r$.
  Finally,
  the discount factor $\gamma$ remains the same.

  We remark that $Q_{M'}^*(s) = V_M^\pi(s)$ for all $s\in \mcal S$ by construction.
  Thus the deterministic (non-replicable) $Q$-function estimation algorithm from \Cref{thm:variance-reduced QVI}
  is in fact a deterministic (non-replicable) algorithm 
  that yields an estimate $\wh V^\pi$ of $V^\pi$
  such that with probability at least $1-\delta$,
  \begin{align*}
    \norm{\wh V^\pi - V^\pi}_\infty &\leq \frac{\varepsilon'}{2N} \\
    \varepsilon' &:= \frac{\varepsilon(\rho - 2\delta)}{\rho+1-2\delta}.
  \end{align*}
  Moreover,
  the sample complexity is
  \[
    \tilde O\left( \frac{\card{\mcal S} N^2}{(1-\gamma)^3 \varepsilon^2 \rho^2} \log\frac1{\delta} \right)
  \]
  since the state-action space of $M'$ has size $\card{\mcal S}$.
  Without loss of accuracy,
  assume we clip the estimates to lie in the feasible range $[0, \nicefrac1{(1-\gamma)}]$.

  In order to replicably estimate $Q^\pi$,
  we use the following relation
  \[
    Q^\pi(s, a) = r(s, a) + \gamma \E_{s'\sim P(\cdot\mid s, a)} \left[ V^\pi(s') \right].
  \]
  Note that we only have access to an estimate $\wh V^\pi(s')$
  and not $V^\pi(s')$,
  thus we instead estimate
  \[
    \bar Q^\pi(s, a) := r(s, a) + \gamma \E_{s'\sim P(\cdot\mid s, a)} \left[ \wh V^\pi(s') \right].
  \]
  But by H\"older's inequality,
  we are still guaranteed that with probability at least $1-\delta$,
  \begin{align*}
    \abs{\bar Q^\pi(s, a) - Q^\pi(s, a)}
    &\leq \gamma \norm{P(\cdot\mid s, a)}_1\cdot \norm{\wh V^\pi - V^\pi}_\infty \\
    &= \gamma \norm{\wh V^\pi - V^\pi}_\infty \\
    &\leq \frac{\varepsilon'}{2N}
  \end{align*}
  for all $s\in \mcal S, a\in \mcal A^s$.

  Each expectation of the form $\E_{s'\sim P(\cdot\mid s, a)} \left[ \wh V^\pi(s') \right]$
  is simply a bounded statistical query.
    By an Hoeffding bound,
    drawing $m$ samples from the generative model $s_j'\sim P(\cdot\mid s, a)$ implies
    \[
        \P\set*{\abs*{\frac1m \sum_{j=1}^m \wh V^\pi(s_j') - \E_{s'} \left[ \wh V^\pi(s') \right]} > \frac{\varepsilon'}{2N}}
        \leq 2\exp\left( -\frac{2m(1-\gamma)^2\varepsilon'^2}{4N^2} \right).
    \]
    Thus with
    \[
        m
        = \tilde O\left( \frac{N^2}{(1-\gamma)^2\varepsilon'^2} \ln\frac1\delta \right)
    \]
    trials,
    the empirical average estimates a single query $\E_{s'\sim P(\cdot\mid s, a)} \left[ \wh V^\pi(s') \right]$
    to an accuracy of $\nicefrac{\varepsilon'}{2N}$
    with probability at least $1-\delta$.
    The total number of samples is thus $Nm$.

  Combining the $V^\pi$ estimation and Hoeffding bound yields an estimate $\bar Q^\pi$ such that
  \[
    \norm{\bar Q^\pi - Q^\pi}\leq \frac{\varepsilon'}{N}
  \]
  with probability at least $1-2\delta$.
  Thus we can use the locally random $\rho$-replicable rounding scheme from \Cref{thm:replicable rounding}
  to compute an estimate $\wh Q^\pi$
  such that with probability at least $1-2\delta$,
  \[
    \norm{\wh Q^\pi - Q^\pi}_\infty \leq \varepsilon.
  \]

  All in all,
  we described an algorithm that replicably estimates $Q^\pi$ given a policy.
  The algorithm is locally random since the only internal randomness used to estimate $Q^\pi(s, a)$
  occurs at the last locally replicable rounding step.
  Finally,
  the total sample complexity of this algorithm is
  \[
    \tilde O\left( \left( \frac{\card{\mcal S} N^2}{(1-\gamma)^3 \varepsilon^2 \rho^2}
    + \frac{N^3}{(1-\gamma)^2 \varepsilon^2 \rho^2} \right)\log\frac1\delta \right)
  \]
  as desired.
\end{pf}

With \Cref{lem:replicable policy to Q} in hand,
we are now ready to prove \Cref{thm:replicable policy lower bound},
which reduces replicable policy estimation to replicable $Q$-function estimation.
We restate the result below for convenience.
\rPolicyLowerBound*

\begin{pf}[\Cref{thm:replicable policy lower bound}]
  Run $\mscr A$ to output a $\rho$-replicable policy $\pi$
  such that
  \[
    \abs{V^\pi(s) - V^\star(s)} \leq \varepsilon
  \]
  with probability at least $1-\nicefrac\delta{12}$
  for each $s\in \mcal S$.

  Applying the algorithm from \Cref{lem:replicable policy to Q}
  with replicability parameter $\rho$,
  failure rate $\nicefrac\delta{12}$,
  and $\pi$ as input yields some estimate $\wh Q^\pi$
  such that
  \[
    \abs*{\wh Q^\pi(s, a) - Q^\pi(s, a)} \leq \varepsilon
  \]
  with probability at least $1-\nicefrac\delta{12}$
  for each $s\in \mcal S, a\in \mcal A^s$.

  Using the relationship
  \[
    Q^\pi(s, a) = r(s, a) + \gamma \E_{s'\sim P(\cdot \mid s, a)} \left[ V^\pi(s) \right],
  \]
  the triangle inequality,
  as well as a union bound,
  we realize that
  \[
    \abs*{\wh Q^\pi(s, a) - Q^\star(s, a)} \leq 2\varepsilon
  \]
  with probability at least $1-\nicefrac\delta{6}$
  for each $s\in \mcal S, a\in \mcal A^s$.
  Moreover,
  $\pi$ is identical across two executions with probability at least $1-\rho$ by assumption
  and thus $\wh Q^\pi$ will be identical across two executions with probability at least $1-2\rho$.

  Remark that the scheme we derived above is a locally random $2\rho$-replicable algorithm for $Q$-estimation.
  It is locally random since the only source of internal randomness
  comes from the locally random subroutine in \Cref{lem:replicable policy to Q}.
  Thus the algorithm satisfies the conditions of \Cref{thm:replicable Q estimation lower bound}
  and the lower bound from that theorem applies.
  In particular,
  if $m$ denotes the sample complexity of $\mscr A$,
  then
  \[
    m + \tilde o\left( \frac{N^3}{(1-\gamma)^3 \varepsilon^2 \rho^2} \right)
    \geq \Omega\left( \frac{N^3}{(1-\gamma)^3 \varepsilon^2 \rho^2} \right).
  \]
  The LHS is the sample complexity of the scheme we derived courtesy of \Cref{lem:replicable policy to Q}
  and the RHS is the lower bound from \Cref{thm:replicable Q estimation lower bound}.
  It follows that
  \[
    m \geq \Omega\left( \frac{N^3}{(1-\gamma)^3 \varepsilon^2 \rho^2} \right)
  \]
  as desired.
\end{pf}

\section{Omitted Details from \texorpdfstring{\Cref{sec:tv ind section}}{TV Indistinguishable Policy Estimation}}\label{apx:tv ind apx}
We now restate and prove \Cref{thm:tv ind oracle for multiple queries},
a $\rho$-TV indistinguishable SQ oracle for multiple queries.
\tvMultipleSQ*

\begin{pf}[\Cref{thm:tv ind oracle for multiple queries}]
    We first argue about the correctness of our algorithm. By assumption,
    with probability at least $1-\nicefrac\delta2$,
    the oracles provide estimates $\wh{\mu}_i$'s such that
    \[
        \max_{i \in [d]} \abs*{\widehat{\mu}_i - v_i} 
        \leq \frac{\varepsilon \rho}{2\sqrt{8d\cdot\log(4d/\delta)}} 
        \leq \frac\varepsilon2 \,.
    \]
    We call this event $\mcal E_1$.
    Fix some $i \in [d]$ and consider the second step where we add Gaussian 
    noise $\mcal N(0,\varepsilon^2)$ to the estimate $\widehat{\mu}_i$
    to obtain noisy estimates $\wh v_i$'s. 
    From standard
    concentration bounds, we know that for $X \sim \mcal N(\mu, \sigma^2), r > 0$
    \[
        \P \set*{\abs{X - \mu} > r \sigma} \leq 2 e^{-r^2/2} \,.
    \]
    Plugging in the values 
    \begin{align*}
        \sigma^2 = \frac{\varepsilon^2}{8\log(4d/\delta))},
        &&
        r = \sqrt{2\log\left( \frac{4d}\delta \right)},
    \end{align*}
    we have that
    \[
        \P \set*{\abs*{\widehat{v}_i - \widehat{\mu}_i} >\varepsilon/2} \leq \frac\delta{2d} \,.
    \]
    Thus, taking a union bound over $i \in [d]$ we have that
    \[
        \P \set*{\max_{i \in [d]}\abs*{\widehat{v}_i - \widehat{\mu}_i} > \varepsilon/2} > \frac\delta2 \,.
    \]
    We call this event $\mcal E_2.$ By taking a union bound over $\mcal E_1, \mcal E_2$
    and applying the triangle inequality,
    we see that
    \[
        \P \set*{\max_{i \in [d]}\abs*{\widehat{v}_i - v_i} > \varepsilon} \leq \delta \,.
    \]

    We now move on to prove the $\rho$-$\TV$ indistinguishability guarantees. Consider two
    executions of the algorithm
    and let $\widehat{\mu}^{(1)}, \widehat{\mu}^{(2)} \in \R^d$ be the estimates
    after the first step of the algorithm.
    We know that, with probability at least $1-\delta$ over the calls to the SQ oracle
    across the two executions it holds that
    \[
        \norm{\widehat{\mu}_1 - \widehat{\mu}_2}_{\infty} 
        \leq \frac{\varepsilon \rho}{\sqrt{8d\cdot\log(4d/\delta)}}  \,.
    \]
    We call this event $\mcal E_3$ and we condition on it for the rest of the proof.
    Standard computations \citep{gupta2020kl}
    reveal that the KL-divergence between two $d$-dimensional Gaussians $p = \mcal N(\mu_p, \Sigma_p), q = \mcal N(\mu_q, \Sigma_q)$ can be written as
    \[
        D_{\mathrm{KL}}(p\|q) = \frac{1}{2}\left(\log\frac{\abs{\Sigma_q}}{\abs{\Sigma_p}} - d + \left(\mu_p - \mu_q\right)\Sigma^{-1}_q\left(\mu_p - \mu_q\right) + \mathrm{trace}\left(\Sigma_q^{-1}\Sigma_p\right)\right)
    \]
    In our setting we have
    \begin{align*}
        p = \mcal N\left( \widehat{\mu}^{(1)}, \frac{\varepsilon^2}{8\log(4d/\delta)}\cdot I_d \right),
        &&
        q = \mcal N\left( \widehat{\mu}^{(2)}, \frac{\varepsilon^2}{8\log(4d/\delta)}\cdot I_d \right).
    \end{align*}
    Plugging
    these in we have that
    \[
        D_{\mathrm{KL}}(p\|q) = \frac{8\log(4d/\delta)}{\varepsilon^2} \norm{\widehat{\mu}^{(1)}-\widehat{\mu}^{(2)}}_2^2 \leq \rho^2 \,.
    \]
    Thus, using Pinsker's inequality we see that
    \[
        \dtv(p,q) \leq \frac{\rho}{\sqrt{2}}
    \]

    Hence, we can bound the expected $\TV$-distance as
    \[
        \E[\dtv(p,q)] \leq \frac{\rho}{\sqrt{2}} + \delta < \rho \,.
    \]
\end{pf}

Next,
we prove \Cref{thm:improved replicable oracle for multiple queries},
an implementation of the $\rho$-TV indistinguishable SQ oracle for multiple queries
whose internal randomness is designed in a way that also ensures $\rho$-replicability.
The result is restated below for convenience.
\rSQByPPP*

\begin{pf}[\Cref{thm:improved replicable oracle for multiple queries}]
    Consider a call to \Cref{alg:tv ind for multiple queries} with TV indistinguishability
    parameter $\nicefrac\rho2.$ Let us call this algorithm $\mscr A$. 
    Notice that given a particular sample,
    $\mscr A$ is Gaussian and therefore absolutely continuous with respect to the Lebesgue measure. 
    Let $\mcal P$
    be the Lebesgue measure over $\R^d.$
    Let $\mcal R$ be a Poisson point process with intensity
    $\mcal P \times \mathrm{Leb} \times \mathrm{Leb}$, 
    where $\mathrm{Leb}$ is the Lebesgue measure over $\R_+$ (cf. \Cref{thm:pairwise opt coupling protocol}). 
    
    The learning rule $\mscr A'$
    is defined as in \Cref{alg:pairwise optimal coupling}. 
    For every input $S$ of $\mscr A$,
    let $f_S$ be the conditional density of $\mscr A$ given the input $S$, let
    $r := \{(x_i,y_i,t_i)\}_{i \in \N}$ be an infinite sequence
    of the Poisson point process $\mcal R$,
    and let $i^\star := \arg\min_{i \in \N}\{t_i: f_S(x_i) > y_i\}$. 
    In words, we consider the pdf of the output conditioned on the input,
    an infinite sequence drawn from the described Poisson point process,
    and we focus on the points of this sequence whose $y$-coordinate
    falls below the curve.
    The output of $\mscr A'$
    is $x_{i^\star}$ and we denote it by $\mscr A'(S;r)$.

    We will shortly explain why this is well-defined, except for a measure zero event.
    The fact
    that $\mscr A'$ is equivalent to $\mscr A$ follows from the coupling guarantees
    of this process
    (cf. \Cref{thm:pairwise opt coupling protocol}),
    applied to the single random vector $\mscr A(S)$.
    We can now observe that,
    except for a measure zero event,
    \begin{enumerate*}[(i)]
        \item since $\mscr A$ is absolutely continuous with respect to $\mcal P$, there exists such a density $f_S$,
        \item the set over which we are taking the minimum is not empty, 
        \item the minimum is attained at a unique point.
    \end{enumerate*}
    This means that $\mscr A'$ is well-defined, 
    except on an event of measure zero\footnote{Under the measure zero event that at least one of these three conditions does not hold, we let $\mscr A'(S;r)$ be some arbitrary $d$-dimensional vector in $[0,1]^d$.}, 
    and by the correctness of the rejection sampling process \citep{angel2019pairwise}, 
    $\mscr A'(S)$ has the desired probability distribution.

    We now prove that $\mscr A'$ is replicable. Since $\mscr A$ is $\nicefrac\rho2$-$\TV$ indistinguishable, it follows that 
    \[\E_{S,S'}[\dtv(\mscr A(S), \mscr A(S'))] \leq \rho/2.\]
    We have shown that $\mscr A'$ is equivalent to $\mscr A$, so we can see that $\E_{S,S'}[\dtv(\mscr A'(S),\mscr A'(S'))] \leq \nicefrac\rho2$. Thus, using the guarantees of \Cref{thm:pairwise opt coupling protocol},
    we have that for any datasets $S,S'$
    \[
        \P_{r \sim \mcal R}\set{\mscr A'(S;r) \neq \mscr A'(S';r)}
        \leq
        \frac{2\dtv(\mscr A'(S),\mscr A'(S'))}{1 + \dtv(\mscr A'(S),\mscr A'(S'))}\,.
    \]
    By taking the expectation over $S,S'$, we get that
    \begin{align*}
        \E_{S,S'}\left[ \P_{r \sim \mcal R}\set{\mscr A'(S,r) \neq \mscr A'(S',r)} \right] &\leq
        \E_{S,S'}\left[ \frac{2\dtv(\mscr A'(S),\mscr A'(S'))}{1 + \dtv(\mscr A'(S),\mscr A'(S'))} \right] \\
        &\leq \frac{2 \E_{S,S'}\left[ \dtv(\mscr A'(S),\mscr A'(S')) \right]}{1 + \E_{S,S' }\left[ \dtv(\mscr A'(S),\mscr A'(S')) \right]} \\
        &\leq \frac{\rho}{1 + \rho/2} \\
        &\leq \rho \,,
    \end{align*}
    where the first inequality follows from \Cref{thm:pairwise opt coupling protocol} and taking the
    expectation over $S, S'$, the second inequality follows from Jensen's inequality, and
    the third inequality follows from the fact that $f(x) = 2x/(1+x)$ is increasing. Now notice that 
    since the source of randomness $\mcal R$ is independent 
    of $S,S'$, we have that
    \[
         \E_{S,S' }\left[\P_{r \sim \mcal R}\set{\mscr A'(S;r) \neq \mscr A'(S';r)} \right] 
         = \P_{S,S', r \sim \mcal R} \set{\mscr A'(S;r) \neq \mscr A'(S';r)} \,.
    \]
    Thus, we have shown that
    \[
        \P_{S,S', r \sim \mcal R} \set{\mscr A'(S;r) \neq \mscr A'(S';r)}
        \leq \rho \,,
    \]
    and so the algorithm $\mscr A'$ is $\rho$-replicable,
    concluding the proof.
\end{pf}

\begin{remark}
    The coupling we described in the previous proof is not 
    computable, since it requires an infinite sequence of samples. 
    However, we can execute it approximately in the following way. We can first
    truncate the tails of the distribution and consider a large enough
    $d$-dimensional box that encloses the pdf of the truncated distribution. 
    Then, by sampling $\widetilde{O}(\exp(d))$ many points we can guarantee
    that, with high probability, there will be at least one that
    falls below the pdf. Even though this approximate coupling is computable,
    it still requires exponential time.
\end{remark}

\begin{remark}[Coordinate-Wise Coupling]\label{rem:coordinate wise coupling 
of Gaussian mechanism}
    Since our algorithms add independent Gaussian noise to each
    of the estimates, a first approach to achieve the coupling using
    only shared randomness would be to construct a pairwise coupling
    between each estimate. In the context of multiple statistical
    query estimation, this would mean that we couple
    the estimate of the $i$-th query in the first execution with 
    the estimate of the $i$-th query in the second execution. Unfortunately,
    even though this coupling is computationally efficient to implement it
    does not give us the desired sample complexity guarantees. To see that,
    notice that when the $\TV$-distance of estimates across each coordinate
    is $O(\rho)$, under this pairwise coupling the probability that at least 
    one of the estimates will be different across the two executions is
    $O(d\cdot\rho)$. However, the $\TV$ distance of the $d$-dimensional
    Gaussians is $O(\sqrt{d} \cdot \rho)$, and this is the reason
    why the more complicated coupling we propose achieves better
    sample complexity guarantees. Our results reaffirms the observation
    that was made by \citet{kalavasis2023statistical} that the 
    replicability property and the sample complexity of an algorithm
    are heavily tied to the implementation of its internal randomness, 
    which can lead to a substantial computational overhead.
\end{remark}

\section{Omitted Details from \texorpdfstring{\Cref{sec:approximately replicable policy estimation}}{Approximately Replicable Policy Estimation}}\label{apx:approximately replicable policy}
We now fill in the missing technical details from \Cref{sec:approximately replicable policy estimation},
which proposes an approximately replicable algorithm for optimal policy estimation.

\begin{algorithm}[H]
\caption{Approximately Replicable Policy Estimation}\label{alg:approximately replicable policy estimation}
\begin{algorithmic}[1]
  \STATE $\delta \gets \min\{\delta, \rho_2/2\}$.
  \STATE Run Sublinear Randomized QVI \citep{sidford2018near} with
   confidence $\delta$, discount factor $\gamma$, and error
   $\rho_1\cdot\varepsilon\cdot(1-\gamma)/(8\log(|\mcal A|)).$
   \STATE Let $\wh Q$ be the output of the previous step.
   \STATE $\lambda \gets \frac{\log(|\mcal A|)}{\varepsilon/2 \cdot (1-\gamma)}$.
   \STATE For every $s\in \mcal S$, let $\pi(s,a) =  \frac{\exp{\lambda \wh Q(s,a)}}{\sum_{a' \in \mcal A} \exp{\lambda \wh Q(s,a')}} .$
   \STATE Output $\pi$.
\end{algorithmic}
\end{algorithm}

\subsection{Different Soft-Max Rules}\label{apx:different
soft-max rules}
Instead of using the exponential soft-max rule
we described in~\Cref{sec:approximately replicable policy estimation}, 
one could use a more sophisticated soft-max rule like the piecewise-linear
soft-max rule that was developed in \citet{epasto2020optimal}. The 
main advantage of this approach is that it achieves a \emph{worst-case}
$\varepsilon$-approximation, i.e., it never picks an action
that will lead to a cumulative reward that is $\varepsilon$ worse
that the optimal one (cf. \Cref{def:soft-max approx}). We also remark
that this leads to sparse policies when there is only a small
number of near-optimal actions.

\subsection{Deferred Proofs}\label{apx:omitted proofs approximately replicable sections}

\subsubsection{Proof of \texorpdfstring{\Cref{lem:exp-soft-max-policy}}{Lemma}}
We now restate and prove \Cref{lem:exp-soft-max-policy},
a useful result which quantifies the performance of the soft-max policy in comparison to the optimal policy.
\arSoftmaxPolicy*

\begin{pf}[\Cref{lem:exp-soft-max-policy}]
    % Let $\wh V(s) = \max_{a \in \mcal A} \wh Q(s, a), \forall s \in \mcal S$.
    % Let $\bar s \in \arg\max_{s \in \mcal S} V^\star(s) - 
    % V^{\wh \pi}(s).$ 
    Using the guarantees of \Cref{lem:exp-soft-max-approx} we have that
    \[
      \max_{a \in \mcal A} \wh Q(s, a) \leq \sum_{a \in \mcal A}\wh \pi(\bar s, a) \cdot \wh Q(s, a) + \varepsilon_2 \,.
    \]
    We can assume without loss of generality that
    $\pi^\star$ is a deterministic policy due to the fundamental theorem
    of RL. 
    Fix an arbitrary $s\in \mcal S$.
    By the fact that $\max_{s \in \mcal S, a \in \mcal A}|\wh Q(s,a) - Q^\star(s,a)| \leq \varepsilon_1$,
    we see that
    \begin{align*}
        Q^\star(s, \pi(s)) &\leq \max_{a \in \mcal A} \wh Q(s, a) + \varepsilon_1 \\
        & \leq \sum_{a \in \mcal A}\wh \pi(\bar s, a) \cdot \wh Q(s, a) + \varepsilon_1 + \varepsilon_2 \\
        & \leq \sum_{a \in \mcal A}\wh \pi(\bar s, a) \cdot  Q^\star(s, a) + 2\varepsilon_1 + \varepsilon_2
    \end{align*}
    An equivalent way to write the previous inequality is
    \begin{align*}
        &r(s,\pi^\star(s)) + \gamma \sum_{s' \in \mcal S}V^\star(s')\cdot P_M(s'|s,\pi^\star(s)) \\
        &\leq
        \sum_{a \in \mcal A} \wh \pi(s, a) \cdot \left( 
        r(s,a)  + \gamma \sum_{s' \in \mcal S}V^\star(s')\cdot P_M(s'|s,a) \right)
        + 2\varepsilon_1 + \varepsilon_2
    \end{align*}
    which implies that
    \begin{align}\label{eq:bounds-rewards-opt-softmax}
    &r(s,\pi^\star(s)) -  \sum_{a \in \mcal A} \wh{\pi}(s, a) \cdot r(s,a) \notag \\
    &\leq \gamma \left( \sum_{a \in \mcal A} \wh \pi(s, a) \cdot \left(\sum_{s' \in \mcal S}V^\star(s')\cdot P_M(s'|s,a)\right) - \sum_{s' \in \mcal S}V^\star(s')\cdot P_M(s'|s,\pi^\star(s))\right) \notag \\
    &\quad + 2\varepsilon_1 + \varepsilon_2 \,.
    \end{align}
        
    Let $\bar s \in \arg\max_{s \in \mcal S} V^\star(s) - 
    V^{\wh \pi}(s).$ Then, we have that
    \begin{align*}
        &V^\star(\bar s) - V^{\wh \pi}(\bar s) \\
        &\leq r(\bar s,\pi^\star(\bar s)) -  \sum_{a \in \mcal A} \wh \pi(\bar s, a) \cdot r(\bar s,a) \\
        &\quad + \gamma\left(\sum_{s' \in \mcal S}V^\star(s')\cdot P_M(s'|\bar s,\pi^\star(\bar s)) - \sum_{a \in \mcal A}\wh \pi(s, a) \cdot \left(\sum_{s' \in \mcal S}V^{\wh \pi}(s')\cdot P_M(s'|\bar s,a)\right) \right) \,.
    \end{align*}
    Bounding the first two terms of the RHS using \Cref{eq:bounds-rewards-opt-softmax} for $s = \bar s$ we see that
    \begin{align*}
        &V^\star(\bar s) - V^{\wh \pi}(\bar s) \\
        &\leq \gamma \left( \sum_{a \in \mcal A} \wh \pi(\bar s, a) \cdot \left(\sum_{s' \in \mcal S}V^\star(s')\cdot P_M(s'|\bar s,a)\right) - \sum_{s' \in \mcal S}V^\star(s')\cdot P_M(s'|\bar s,\pi^\star(\bar s))\right) \\
        &\quad + 
        \gamma\left(\sum_{s' \in \mcal S}V^\star(s')\cdot P_M(s'|\bar s,\pi^\star(\bar s)) - \sum_{a \in \mcal A}\wh \pi(\bar s, a) \cdot \left(\sum_{s' \in \mcal S}V^{\wh \pi}(s')\cdot P_M(s'|\bar s,a)\right) \right) \\
        &\quad + 2\varepsilon_1 + \varepsilon_2 \,.
    \end{align*}
    After we cancel out some terms on the RHS we conclude that
    \begin{align*}
         &V^\star(\bar s) - V^{\wh \pi}(\bar s) \\
         &\leq \gamma\left( \sum_{a \in \mcal A} \wh \pi(\bar s, a) \cdot \left(\sum_{s' \in \mcal S}\left(V^\star(s') - V^{\wh \pi}(s')\right)\cdot P_M(s'|\bar s,a)\right) \right) + 2\varepsilon_1 + \varepsilon_2 \\
         & \leq \gamma\left( \sum_{a \in \mcal A} \wh \pi(\bar s, a) \cdot \left(\sum_{s' \in \mcal S}\left(V^\star(\bar s) - V^{\wh \pi}(\bar s)\right)\cdot P_M(s'|\bar s,a)\right) \right) + 2\varepsilon_1 + \varepsilon_2\\
         & = \gamma\left(V^\star(\bar s) - V^{\wh \pi}(\bar s) \right)   + 2\varepsilon_1 + \varepsilon_2 \,.
    \end{align*}
    The second inequality follows from the fact that $\bar s \in \arg\max_{s \in \mcal S} V^\star(s) - V^{\wh \pi}(s)$ and the equality follows from the fact that $\sum_{a \in \mcal A}\wh \pi(\bar s, a) = 1, \sum_{s' \in \mcal S}P_M(s'|\bar s, a) = 1, \forall a \in \mcal A.$
    Rearranging, we see that
    \begin{align*}
        V^\star(\bar s) - V^{\wh \pi}(\bar s) \leq \frac{2\varepsilon_1 + \varepsilon_2}{1-\gamma} \,.
    \end{align*}
    But the choice of $s\in \mcal S$ was arbitrary,
    concluding the proof.
    \iffalse
    Hence, 
    \begin{align*}
        V^\star(s) - V^{\wh \pi}(s) &\leq \frac{2\varepsilon_1 + \varepsilon_2}{1-\gamma},\quad\forall s \in \mcal S \,. \qedhere
    \end{align*}
    \fi
\end{pf}

\subsubsection{Proof of \texorpdfstring{\Cref{thm:approximately-replicable-policy-estimation-sample
-complexity}}{Theorem}}
Last but not least,
we prove \Cref{thm:approximately-replicable-policy-estimation-sample
-complexity},
an upper bound on approximately-replicable policy estimation.
We restate the result below for convenience.
\arPolicyEstimation*

\begin{pf}[\Cref{thm:approximately-replicable-policy-estimation-sample
-complexity}]
    First, we argue about the correctness of the approach. Using \Cref{thm:variance-reduced QVI} with parameters
    $\gamma, \delta/2, \varepsilon_1 = \rho_1\cdot\varepsilon\cdot(1-\gamma)/(8\log(|\mcal A|))$, we can estimate some
    $\wh Q_1$ such that $||\wh Q_1 - Q^\star||_\infty \leq \varepsilon_1$, with probability at least $1-\delta/2.$ We call this event
    $\mcal E_1$ and we condition on it. Since we use the exponential
    mechanism with parameter $\lambda = \log(|\mcal A|)/(\varepsilon/2 \cdot (1-\gamma))$ to compute a policy $\wh \pi(s)$, for every state $s \in \mcal S$, \Cref{lem:exp-soft-max-policy} guarantees that 
    \[
        ||V^\star - V^{\wh \pi}||_\infty \leq \frac{2\rho_1\cdot\varepsilon\cdot(1-\gamma)/(8 \log(|\mcal A|)) + \varepsilon/2\cdot(1-\gamma)}{1-\gamma} \leq \varepsilon \,.
    \]
    This completes the proof of correctness.

    We now proceed with showing the replicability guarantees of this process. Consider two executions of the algorithm and let $\wh Q_1, 
    \wh Q_2$ denote the output of the algorithm described in \Cref{thm:variance-reduced QVI} in the first and second run, respectively. Notice that, with probability at least $1-\min\{\delta,\rho_2\}$
    it holds that $||\wh Q_1 - Q^\star||_\infty \leq \varepsilon_1, ||\wh Q_2 - Q^\star||_\infty \leq \varepsilon_1$. We call this event
    $\mcal E_2$ and condition on it for the rest of the proof. Thus,
    by the triangle inequality, 
    we have that
    \[
        ||\wh Q_1 - \wh Q_2||_\infty \leq 2\varepsilon_1 \,.
    \]
    Let $\wh \pi_1, \wh \pi_2$ denote the policy the algorithm outputs
    in the first and second execution, respectively. Let $s \in \mcal S$
    be some arbitrary state. Since the exponential soft-max
    is $2\lambda$-Lipschitz continuous with respect to $(\ell_\infty, D_\alpha)$ we have that
    \begin{align*}
        D_\alpha(\pi_1(s)||\pi_2(s)) &\leq 2\lambda ||\wh Q_1 - \wh Q_2||_\infty \\
        &= 2\cdot \frac{\log(|\mcal A|)}{(\varepsilon/2 \cdot (1-\gamma))} \cdot 2\cdot \frac{\rho_1\cdot\varepsilon\cdot(1-\gamma)}{(8\log(|\mcal A|))} \\
        & = \rho_1 \,.
    \end{align*}
    Since the event $\mcal E_2$ happens with probability $1-\rho_2$
    we see that the algorithm is $(\rho_1,\rho_2)$-approximately replicable
    with respect to $D_\alpha$, i.e., the Renyi divergence of order $\alpha$. This concludes the approximate replicability part of the proof.

    As a last step, we bound the sample complexity of our algorithm. The 
    stated bound follows directly from \Cref{thm:variance-reduced QVI}
    since we call this algorithm with error parameter $ \rho_1\cdot\varepsilon\cdot(1-\gamma)/(8\log(|\mcal A|)).$
\end{pf}

 \begin{remark}[Replicability Under TV Distance]
It is known that the TV distance of two probability
distributions is upper bounded by $D_\infty.$ Thus,
we can see that \Cref{thm:approximately-replicable-policy-estimation-sample
-complexity} provides the same guarantees when we want to
establish replicability with respect to the TV distance.
\end{remark}

\begin{remark}[Sample Complexity Dependence on Replicability Parameters]
Notice that the dependence of the number of samples in \Cref{thm:approximately-replicable-policy-estimation-sample
-complexity}
on the two different replicability parameters of \Cref{def:approx-replicable-policy} is different. In particular, the dependence on $\rho_2$ is $\polylog(\nicefrac1{\rho_2})$, whereas the dependence on $\rho_1$ is
$\poly(\nicefrac1{\rho_1})$. 
\end{remark}

\section{Guarantees under Different Replicability Notions}
\label{sec:discussion regarding different repl notions}
Since we have studied three different replicability notions in this work,
we believe it is informative to discuss the advantages and the drawbacks of 
each one of them. Our discussion is centered across four different axes:
the replicability guarantees that each notion provides, 
the sample complexity required to satisfy each definition, the running
time required to run the underlying algorithms, and the ability 
to test/verify whether the algorithms have the desired replicability properties.

The definition of \cite{impagliazzo2022reproducibility} (\Cref{def:replicability})
provides the strongest replicability guarantees since it requires that the two
outputs are exactly the same across the two executions. It is also computationally
efficient to verify it. Even though it is statistically equivalent to the definition
of $\TV$ indistinguishability, our results along with the results of
\citet{bun2023stability, kalavasis2023statistical} indicate that there might be a 
computational separation between these two notions. Moreover, the fact that this notion
is so tightly related to the way the internal randomness of the algorithm is implemented
is a property that is not exhibited by any other notion of stability we are aware of
and can be problematic in some applications. 

The definition of TV indistinguishability of \citet{kalavasis2023statistical} (\Cref{def:tv ind})
provides strong replicability guarantees, in the sense that someone who observes
the outputs of the algorithm under two executions when the inputs are $S, S'$, cannot
distinguish which one between $S,S'$ was responsible for generating this output. 
Moreover, this definition does \emph{not} depend on the way the internal randomness 
of the algorithm is implemented. On the other hand, testing whether an algorithm
has this property is more subtle compared to \Cref{def:replicability}. In the case
of the Gaussian mechanism-based algorithms we discuss in this work the following holds:
if the output of the algorithm is \emph{promised} to be drawn from a Gaussian distribution,
it is computationally and statistically efficient to test whether the outputs
under two different datasets $S, S'$ are close in $\TV$ distance. However, it is not 
clear how one can test if the outputs are indeed drawn from a Gaussian distribution.

Finally, the notion of approximate replicability (\Cref{def:approximate replicability})
we introduce is a further relaxation 
of the TV indistinguishability property in the following sense: both the replicability 
definition and TV indistinguishability definition treat the outputs in a ``binary'' manner, in the
sense that they only care whether the outputs are exactly the same across the two executions.
This definition takes a more nuanced approach and considers some notion of distance 
across the outputs that is not binary. As a result, it provides the weakest
replicability guarantees, which, however, could be sufficient in most RL applications. 
Moreover, as our results indicate,
there might be some inherent advantage in terms of the sample
complexity required to achieve this notion compared to (strict) replicability or
TV indistinguishability, which can be crucial in RL applications with large state-action space.
Moreover, similarly as with the replicability definition, it is 
also efficient to test whether an algorithm has this property or not.

To sum up, even though we have not completely characterized the sample
complexity and computational complexity of each definition we believe that the following 
is the complete picture: the replicability property is statistically equivalent to the TV
indistinguishability property and the approximate replicability property has sample
complexity that is smaller by a factor of $N$. Moreover, we believe
that there is a computational gap between the notions of replicability and TV indistinguishability.
We underline that under \Cref{conj:replicable randomized coin lower bound}, the results
of our work give a complete characterization\footnote{Up to poly-logarithmic factors.}
of the sample complexity of these problems with respect to $N.$ 

\end{document}